\documentclass[10pt,journal,compsoc]{IEEEtran}

\usepackage{amsmath}
\usepackage{amsthm}
\usepackage{graphicx}
\usepackage{amssymb}
\usepackage{cite}
\usepackage[caption=false,font=normalsize,labelfont=sf,textfont=sf]{subfig}
\newtheorem{prop}{Proposition}
\newtheorem{theorem}{Theorem}
\newtheorem{lemma}{Lemma}
\usepackage{xcolor}

\newcommand{\emjw}{e^{-\mathrm{j}\omega}}

\begin{document}
\title{Wavelet Design in a Learning Framework}
	\author{Dhruv~Jawali,~Abhishek~Kumar~and~Chandra~Sekhar~Seelamantula,~\IEEEmembership{Senior Member,~IEEE}
		\IEEEcompsocitemizethanks{
		\IEEEcompsocthanksitem This work has been submitted to the IEEE Transactions on Pattern Analysis and Machine Intelligence for possible publication. Copyright may be transferred without notice, after which this version may no longer be accessible. \protect\\
		\IEEEcompsocthanksitem D. Jawali is with the National Mathematics Initiative, Indian Institute of Science, Bangalore - 560012, India (Email: dhruv13@iisc.ac.in). 
		\IEEEcompsocthanksitem A. Kumar is presently with the Electrical and Computer Engineering Department, Rice University, Texas 77005, USA (Email: ak109@rice.edu).
		\IEEEcompsocthanksitem C. S. Seelamantula is with the Department of Electrical Engineering, Indian Institute of Science, Bangalore - 560012, India (Email: chandra.sekhar@ieee.org, Tel.: +91 80 22932695, Fax: +91 80 23600444). \protect\\
		\IEEEcompsocthanksitem \indent Sections~\ref{sec:fb_ae}, \ref{sec:vm} and \ref{sec:perf_measures}  appeared in the Proceedings of IEEE International Conference on Acoustics, Speech, and Signal Processing 2019 \cite{icassp2019}.}}%
\markboth{}{Wavelet Design in a Learning Framework}
\IEEEtitleabstractindextext{%
\begin{abstract}
Wavelets have proven to be highly successful in several signal and image processing applications. Wavelet design has been an active field of research for over two decades, with the problem often being approached from an analytical perspective. In this paper, we introduce a learning based approach to wavelet design. We draw a parallel between convolutional autoencoders and wavelet multiresolution approximation, and show how the learning angle provides a coherent computational framework for addressing the design problem. We aim at designing data-independent wavelets by training filterbank autoencoders, which precludes the need for customized datasets. In fact, we use high-dimensional Gaussian vectors for training filterbank autoencoders, and show that a near-zero training loss implies that the learnt filters satisfy the perfect reconstruction property with very high probability. Properties of a wavelet such as orthogonality, compact support, smoothness, symmetry, and vanishing moments can be incorporated by designing the autoencoder architecture appropriately and with a suitable regularization term added to the mean-squared error cost used in the learning process. Our approach not only recovers the well known Daubechies family of orthogonal wavelets and the Cohen-Daubechies-Feauveau family of symmetric biorthogonal wavelets, but also learns wavelets outside these families.
\end{abstract}

}

\maketitle

\IEEEdisplaynontitleabstractindextext
\IEEEpeerreviewmaketitle

\IEEEraisesectionheading{\section{Introduction}\label{sec:introduction}}
\label{sec:intro}
    \IEEEPARstart{W}{avelet} representations are at the heart of several successful signal and image processing applications developed over the past three decades.  Wavelets endowed with important properties such as {\it vanishing moments} are ideal for detecting singularities and edges in signals  \cite{mallat1992singularity, mallat1992characterization}. The ability of wavelets to produce sparse representations of natural and medical images has led to several successes in solving inverse problems such as image denoising \cite{donoho1995adapting, luisier2010sure, luisier2012cure, seelamantula2015image}, deconvolution/deblurring \cite{figueiredo2003algorithm, bioucas2007new, li2017pure}, reconstruction \cite{chan2003wavelet}, image compression \cite{taubman2002jpeg2000} etc. Wavelets have also given rise to multidimensional generalizations such as curvelets \cite{candes2004new}, contourlets \cite{do2005contourlet}, surfacelets \cite{lu2007multidimensional}, directionlets \cite{velisavljevic2006directionlets}, etc. A review of these representations can be found in \cite{jacques2011panorama}.\\
	\indent Wavelet filterbanks implementing a multiresolution approximation have been incorporated as fixed/non-trainable blocks within trainable convolutional neural networks \cite{kang2017deep, liu2018multi}, achieving state-of-the-art performance for inverse problems in image restoration. Scattering convolutional networks incorporating wavelets have been developed by Bruna and Mallat \cite{bruna2013invariant}. The resulting representations are translation-invariant, stable to deformations, and preserve high-frequency content, all of which are important requirements for achieving high-accuracy image classification performance. Hybrid architectures that use the wavelet scattering network in conjunction with trainable neural network blocks have been shown to provide competitive image classification performance, and have also been used to represent images within a generative adversarial framework \cite{oyallon2018scattering}. Such phenomenal successes have been possible with wavelets because of their ability to provide parsimonious and invariant representations.\\
    \indent Wavelet design lies at the interface between signal processing and mathematics, and has benefited greatly by a fruitful exchange of ideas between the two communities. Wavelets can be designed to constitute a frame \cite{kovacevic2007lifea, kovacevic2007lifeb}, Riesz basis, or orthonormal basis for the space of square-integrable functions $L^2(\mathbb{R})$ \cite{primer1998introduction, mallat2008wavelet, chui2016introduction}. Of particular interest is the orthonormal flavor in a multiresolution approximation (MRA) setting, which comprises a nested subspace structure that allows one to move seamlessly across various resolutions. In particular, considering the dyadic MRA and orthonormal wavelet bases satisfying the two-scale equation, efficient algorithms have been developed to compute the projections of a function $f \in L^2(\mathbb{R})$ at different resolutions --- this property also establishes a close connection between wavelet decomposition and filterbank analysis \cite{Vetterli86, mallat1989theory}. Thanks to this connection, the problem of designing orthonormal wavelet bases that satisfy chosen properties becomes equivalent to one of optimizing discrete filters that obey certain design conditions. One could, therefore, start with a discrete filter and determine the corresponding wavelet or vice versa \cite{primer1998introduction, mallat2008wavelet}.
    
    \subsection{Motivation for This Paper}
	The design of wavelets has largely been carried out analytically. For instance, consider the construction of Daubechies wavelets of a certain order and vanishing moments \cite{daubechies1988orthonormal, daubechies1992ten}. The vanishing moments property is incorporated explicitly following the Strang-Fix condition \cite{strang2011fourier}, which partly determines the wavelet filter. The remaining part of the filter is determined by enforcing the conjugate-mirror filter condition and solving for the filter coefficients based on Bezout's theorem and Kolmogorov spectral factorization. Filterbank design, on the other hand, employs sophisticated optimization machinery to enforce the perfect reconstruction conditions \cite{vetterli1995wavelets, strang1996wavelets, vaidyanathan2006multirate}.\\
	\indent In this paper, we ask if one could adopt a {\it learning} strategy to revisit the problem of designing perfect reconstruction filterbanks (PRFBs) and wavelets. Given the enormous attention that machine learning approaches have been receiving of late due to their phenomenal successes, the quest is but pertinent. Our objective is to leverage the state-of-the-art machine learning techniques and specialized tools for solving the problem of wavelet design. We draw a parallel between convolutional autoencoders and perfect reconstruction filterbanks, which casts the design problem within a learning framework. Recently, Pfister and Bresler introduced an undecimated filterbank design approach to learning \emph{data-adaptive} sparsifying transforms \cite{pfister2018learning}. They employed techniques such as stochastic gradient descent and automatic differentiation to optimize the filters. The problem of learning data-adaptive wavelet frames for solving inverse problems has also been explored by Tai and E \cite{jmlrwaveletlearning}. In this paper, our objective is to design \emph{data-independent} two-channel critically sampled PRFBs satisfying certain properties, and thereafter, to determine the corresponding wavelet bases using the recently developed tools in machine learning.
	
	\subsection{Our Contributions}
	We begin by reviewing the crucial connections between a dyadic wavelet transform and PRFBs (Section~\ref{sec:overview}), which are well established in the literature and also form the foundation for the viewpoint adopted in this paper. The starting point for our developments is to interpret a PRFB as a convolutional autoencoder but without the max-pool and activation nonlinearities in the encoder/decoder blocks --- essentially, we have a {\it filterbank autoencoder}. Once this analogy is established, a plethora of techniques and tools used in state-of-the-art deep learning approaches become readily applicable. We train the filterbank autoencoder by minimizing the mean-square error (MSE) loss with the training data being high-dimensional Gaussian vectors (Section~\ref{sec:wv_lf}). The use of Gaussian vectors for training serves as a scaffolding to steer the optimization objective and renders the autoencoder data-independent. Properties such as sparsity for piecewise-regular signals have to be incorporated via other means, for instance, using vanishing moments. We then proceed with designing 1-D orthonormal wavelets with a specified number of vanishing moments (Section~\ref{sec:1d_wv}). The formalism is adapted to handle the biorthogonal case as well. The parameters of the training procedure are the filter lengths and the number of vanishing moments. By an appropriate choice of the parameters, we obtain several well-known wavelets, for instance the Daubechies wavelets and Symmlets in the orthonormal case and members of the Cohen-Daubechies-Feauveau (CDF) family \cite{cohen1992biorthogonal} in the birthogonal case. We do impose necessary checks to ensure that the learnt filters indeed generate stable Riesz bases. The proposed framework enables one to learn wavelets that are outside of these classes as well. For designs that have conflicting constraints and for which no solution exists, for instance, perfect symmetry in the case of orthonormal wavelets with more than one vanishing moment, our framework indicates that as well. We also show that one could also design asymmetric biorthogonal wavelets using our approach. Concluding remarks and potential directions for future work are presented in Section~\ref{sec:conclusions}.
	
	\subsection{Notation}
	Scalars are denoted by lowercase letters (e.g. $m$), vectors by boldface lowercase letters (e.g. $\boldsymbol{h}$) and matrices by boldface uppercase letters (e.g. $\boldsymbol{H}$). The transpose of $\boldsymbol{H}$ is denoted as $\boldsymbol{H}^{\mathrm{T}}$. The $n \times n$ identity and zero matrices are denoted by $\boldsymbol{I}_n$ and $\boldsymbol{0}_n$, respectively. The zero vector is denoted by $\boldsymbol{0}$. Discrete signals are represented as $h[n]$. In this paper, we focus only on compactly supported filters. Therefore, interpreting a 1-D filter $h[n]$ as a finite-dimensional vector $\boldsymbol{h}$ is natural. Fourier transforms are denoted by a caret or hat notation. The context would disambiguate whether the Fourier transform is in the continuous-time domain or the discrete-time domain. For instance, given a sequence $h[n]$, the discrete-time Fourier transform (DTFT) is denoted by $\hat{h}(\omega)$, which is $2\pi$-periodic. Similarly, for a function $\phi(t)$, the continuous-time Fourier transform (CTFT) is denoted by $\hat{\phi}(\omega)$, which is in general not periodic. The space of square-integrable functions is denoted as $L^2(\mathbb{R})$, and the corresponding inner-product is defined as $\langle x, y \rangle = \int_{-\infty}^{+\infty} x(t) y^*(t)\,\mathrm{d}t$, where $y^*(t)$ is the complex conjugate of $y(t)$. The space of square-summable sequences is denoted as $\ell^2(\mathbb{Z})$, and the corresponding inner-product is denoted by $\langle x, y \rangle = \displaystyle\sum_{n \in \mathbb{Z}} x[n] y^*[n]$.
    
\section{A Review of 1-D Wavelet Design}
	\label{sec:overview}
\indent A wavelet function $\psi(t) \in L^2(\mathbb{R})$ has zero average, i.e., $\int \psi(t)\,\mathrm{d}t = 0$, and is localized in the time-frequency plane. The family of functions  $$\left\{\psi_{a,b}(t) = \frac{1}{\sqrt{a}}\psi\left(\frac{t-b}{a}\right)\right\}_{a \in \mathbb{R}^+, b \in \mathbb{R}},$$
obtained by time-shifting and scaling the prototype function $\psi(t)$, could be designed to constitute either overcomplete or orthonormal bases for functions in  $L^2(\mathbb{R})$.

\subsection{Multiresolution Approximation}
\indent Central to wavelet analysis is the multiresolution approximation developed by Mallat \cite{mallat1989theory}.
	A multiresolution approximation (cf. Definition 7.1, \cite{mallat2008wavelet}) is a sequence of closed nested subspaces $\{ V_j \}_{j \in \mathbb{Z}}$ of $L^2(\mathbb{R})$ --- called the {\it approximation subspaces} ---  which can be used to represent signals at varying resolutions. The space $V_j$ corresponding to resolution $2^{-j}$ is spanned by the orthonormal bases $\{ \phi_{j,n}(t) := \phi_j(t - n) \}_{n \in \mathbb{Z}}$, where $\phi_j(t)$ is obtained by time-scaling the prototype function $\phi(t)$ as follows:
	$$
	\phi_j(t) = \frac{1}{\sqrt{2^{j}}} \phi \left( \frac{t}{2^{j}} \right).
	$$
	The function $\phi(t)$ is the \emph{scaling function} and constitutes the generator kernel for the space $V_0$. The nesting property means that a lower-resolution subspace $V_{j + 1}$ is contained within a higher-resolution space $V_j$, i.e., $V_{j+1} \subset V_j, \forall j \in \mathbb{Z}$. Let $W_{j+1}$ denote the orthogonal complement of $V_{j+1}$ in $V_j$, which leads to the direct-sum decomposition: $V_{j} = V_{j + 1} \oplus W_{j + 1}$. $W_j$ is the {\it detail subspace} and is spanned by $\{\psi_{j,n}(t) := \psi_j(t - n)\}_{n \in \mathbb{Z}}$, where $\psi_{j,n}(t)$ is the time-scaled and translated version of the \emph{wavelet function} $\psi(t)$. The generators for $V_{j + 1}$ and $W_{j + 1}$ satisfy the \emph{two-scale equations}:
	\begin{flalign}
	\label{eqn:2s_phi}
	\phi_{j + 1} \left( t \right) &= \sum\limits_{n \in \mathbb{Z}} h[n] \phi_j(t - n), \text{and}\\ 
	\label{eqn:2s_psi}
	\psi_{j + 1} \left( t \right) &=  \sum\limits_{n \in \mathbb{Z}} g[n] \phi_j(t - n),
	\end{flalign}
 	respectively, where the filters $h[n]$ and $g[n]$ are referred to as the {\it scaling} and {\it wavelet} filters, respectively. Given a signal $f \in L^2(\mathbb{R})$, the {\it approximation} and {\it detail coefficients} at resolution $2^{-j}$, denoted by $a_j[n]$ and $d_j[n]$, respectively, are given by the orthogonal projection of $f$ onto the spaces $V_j$ and $W_j$, respectively:
	\begin{flalign}
	a_j[n] = \langle f, \phi_{j, n} \rangle; \quad d_j[n] = \langle f, \psi_{j, n} \rangle.
	\end{flalign}
	The representation coefficients at various scales are computed efficiently by Mallat's algorithm \cite{mallat1989theory, mallat2008wavelet}: 
	\begin{flalign}
	\label{eqn:fb_1}
	a_{j + 1}[n] = \sum\limits_{m \in \mathbb{Z}} h[m - 2n] a_j[m] = (a_j * \bar{h})[2n], \\
	\label{eqn:fb_2}
	d_{j + 1}[n] = \sum\limits_{m \in \mathbb{Z}} g[m - 2n] a_j[m] = (a_j * \bar{g})[2n],
	\end{flalign}
	where $\bar{h}[n] := h[-n]$ and $\bar{g}[n] := g[-n]$ are the corresponding time-reversed sequences. Equations (4) and (5) correspond to the lowpass and highpass outputs of the analysis filterbank illustrated in \figurename{~\ref{fig:prfb}}, with input $x[n] = a_j[n]$. The higher-resolution approximation coefficients are computed using the lower-resolution approximation and detail coefficients as follows:
	\begin{align}
	\label{eqn:fb_syn}
	a_{j}[n] = \sum\limits_{m \in \mathbb{Z}} \tilde{h}[n - 2m] a_{j+1}[m] + \sum\limits_{m \in \mathbb{Z}} \tilde{g}[n - 2m] d_{j+1}[m],
	\end{align}
	which corresponds to the output of the synthesis filterbank illustrated in \figurename{~\ref{fig:prfb}}, where the filters are related as $\tilde{h}[n] = h[n]$, $\tilde{g}[n] = g[n]$, and $g[n] = (-1)^{1-n} h[1 - n]$. 
	So far, we have assumed that $\{ \phi_{j, n}(t) \}_{j, n \in \mathbb{Z}}$ and $\{ \psi_{j, n}(t) \}_{j, n \in \mathbb{Z}}$ form orthonormal bases for $V_j$ and $W_j$, respectively. Instead of orthonormal bases, one could also consider Riesz bases for the spaces $V_j$ and $W_j$. A Riesz basis $\{ \phi_{j, n}(t) \}_{j, n \in \mathbb{Z}}$ of the space $V_j$ is characterized by two constants $0 < \sigma_{\text{min}} \leq \sigma_{\text{max}} < \infty$ such that for all $x(t) \in V_j$,
	$$
	    \sigma_{\text{min}} \|x\|_2^2 \leq \sum_{j, n \in \mathbb{Z}} |\langle x, \phi_{j, n} \rangle|^2 \leq \sigma_{\text{max}} \|x\|_2^2.
	$$ 
	The dual spaces $\tilde{V}_j$ and $\tilde{W}_j$ corresponding to the Riesz bases $V_j$ and $W_j$, respectively, would be spanned by their biorthogonal counterparts $\{ \tilde{\phi}_{j^{\prime}, n^{\prime}}(t) \}_{j^{\prime}, n^{\prime} \in \mathbb{Z}}$ and $\{ \tilde{\psi}_{j^{\prime}, n^{\prime}}(t) \}_{j^{\prime}, n^{\prime} \in \mathbb{Z}}$, such that:
	\begin{flalign*}
	    \langle \phi_{j, n}, \tilde{\phi}_{j^{\prime}, n^{\prime}} \rangle =
	    \langle \psi_{j, n}, \tilde{\psi}_{j^{\prime}, n^{\prime}} \rangle = \delta[j - j^{\prime}] . \delta[n - n^{\prime}],
	\end{flalign*}
	where $\delta[\cdot]$ denotes the Kronecker delta. The biorthogonal bases also satisfy two-scale equations, with corresponding scaling and wavelet filters $\tilde{h}[n]$ and $\tilde{g}[n]$, respectively. Just as $h[n]$ and $g[n]$ constitute the lowpass and highpass filters on the analysis side, the filters $\tilde{h}[n]$ and $\tilde{g}[n]$ constitute their counterparts on the synthesis side, resulting in a perfect reconstruction filterbank.

	\subsection{Perfect Reconstruction Filterbanks}
	\begin{figure}[t]
		\centering
		\includegraphics[width=3.2in]{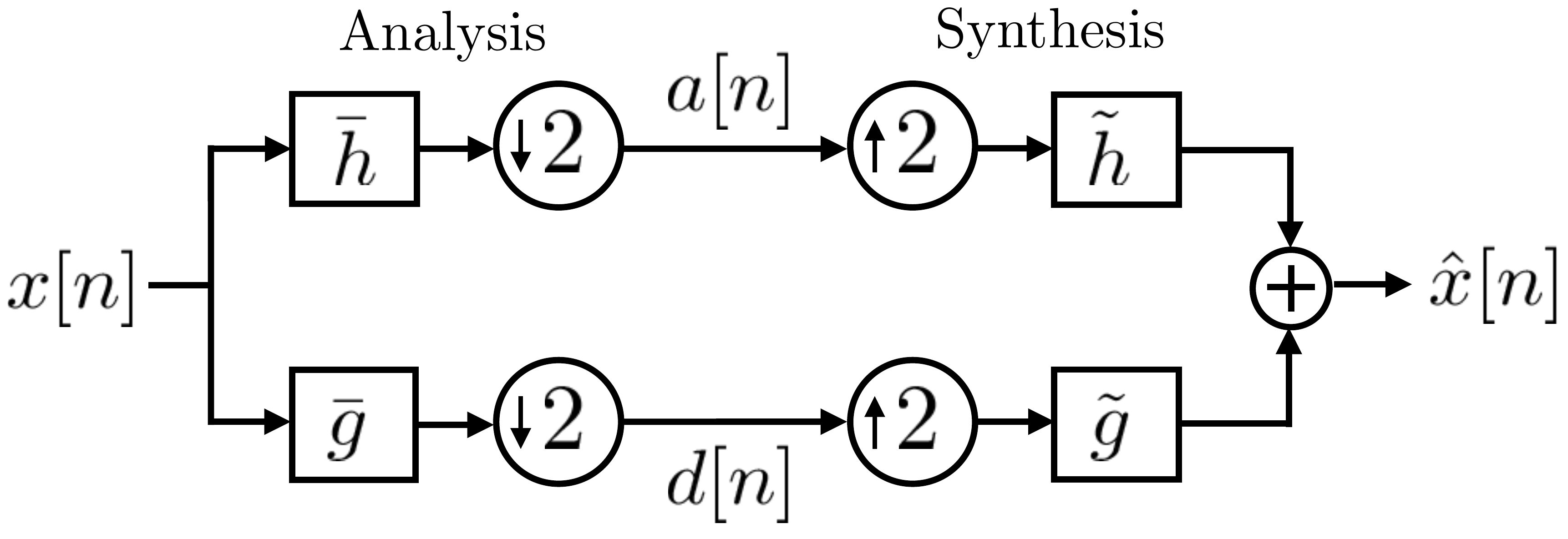}
		\caption{A 1-D two-channel filterbank, with analysis filters $\bar{h}[n], \bar{g}[n]$, and synthesis filters $\tilde{h}[n], \tilde{g}[n]$. The outputs $a[n]$ and $d[n]$ of the analysis filterbank are referred to as the approximation and detail coefficients, respectively.}
		\label{fig:prfb}
	\end{figure}
    Consider the two-channel filterbank shown in \figurename{~\ref{fig:prfb}}. Enforcing ${\hat x}[n] = x[n]$ gives rise to the following perfect reconstruction (PR) conditions \cite{Vetterli86}:
	\begin{alignat}{3}
	\label{eqn:pr}
	& \text{PR-1}: \quad &\hat{h}^*(\omega) \hat{\tilde{h}}(\omega) &+ \hat{g}^*(\omega) \hat{\tilde{g}}(\omega) = 2, \\
	\label{eqn:ac}
	 \text{and}\quad&\text{PR-2}: \quad &\hat{h}^*(\omega + \pi) \hat{\tilde{h}}(\omega) &+ \hat{g}^*(\omega + \pi) \hat{\tilde{g}}(\omega) = 0.
	\end{alignat}
	Constraining the filters $g[n]$ and $\tilde{g}[n]$ in terms of $\tilde{h}[n]$ and $h[n]$ as follows:
	\begin{flalign}
	\label{eqn:ac1}
	g[n] &= a\,(-1)^{1 - n} \tilde{h}[1 - n], \text{and}\\
	\label{eqn:ac2}
	\tilde{g}[n] &= a^{-1} (-1)^{1 - n} h[1 - n],
	\end{flalign}
	ensures that the PR-2 condition in Equation~(\ref{eqn:ac}) is automatically satisfied. The PR-1 condition then takes the form:
	\begin{flalign}
	\label{eqn:pr1}
	\hat{h}^*(\omega) \hat{\tilde{h}}(\omega) &+ \hat{h}^*(\omega + \pi) \hat{\tilde{h}}(\omega + \pi) = 2,
	\end{flalign}
	which is both necessary and sufficient for the filterbank to be PR. Hence, the perfect reconstruction filterbank is completely specified by the two filters $h[n]$ and $\tilde{h}[n]$, and is referred to as a \emph{biorthogonal} filterbank \cite{cvetkovic1998oversampled}. Further, setting $\tilde{h}[n] = h[n]$ reduces it to an \emph{orthogonal} filterbank, which satisfies the conjugate mirror filter (CMF) condition:
	\begin{flalign}
	\label{eqn:qmf}
	|\hat{h}(\omega)|^2 + |\hat{h}(\omega + \pi)|^2 = 2.
	\end{flalign}
	Thus, the multiresolution approximation and PRFBs are very closely related to each other.
	
	\subsection{Wavelet Construction From Orthogonal Filterbanks}
	\begin{figure}[t]
		\centering
		\includegraphics[width=3.2in]{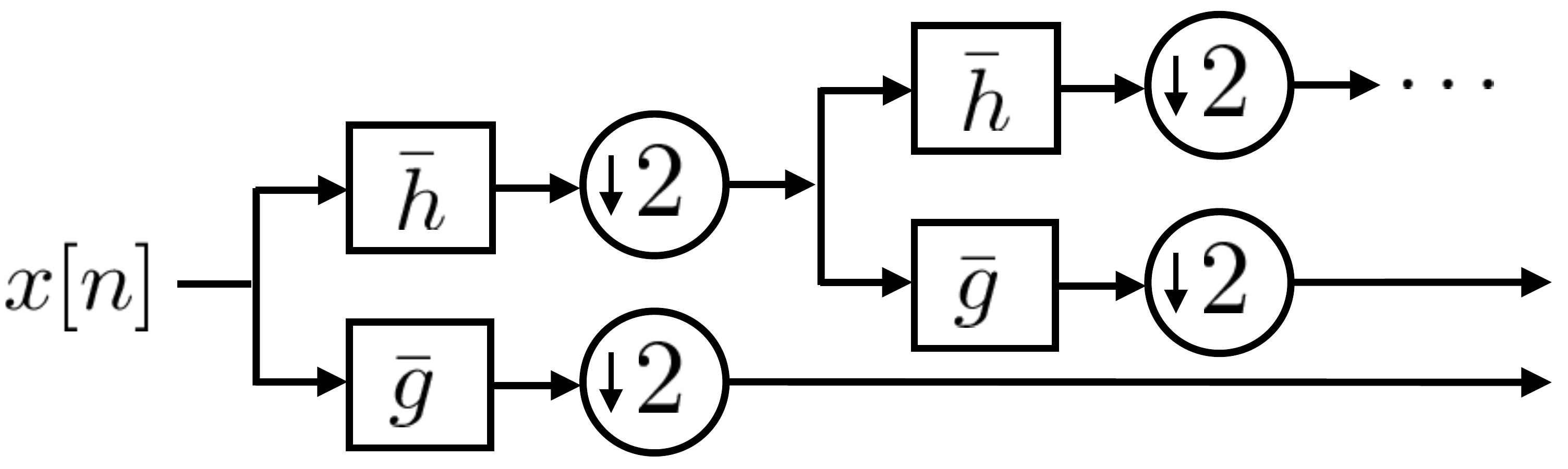}
		\caption{An infinitely cascaded two-channel filterbank, obtained by iterating the analysis filterbank over the approximation coefficients at each level. If the filters $\bar{h}[n]$ and $\bar{g}[n]$ are the scaling and wavelet filters, respectively, then the cascaded filterbank implements the discrete wavelet transform efficiently. An infinite cascade of the scaling filter $\bar{h}[n]$, upon convergence, gives rise to the scaling function $\phi(t)$, and can also be used to obtain the wavelet function $\psi(t)$ using Equation (\ref{eqn:filttowave}).}
		\label{fig:cascade}
	\end{figure}
	A key result by Mallat and Meyer \cite{meyer1986ondelettes, mallat1989theory} states that an orthogonal filterbank gives rise to a wavelet multiresolution approximation if, and only if, the filter $h[n]$ satisfying (\ref{eqn:qmf}) also satisfies $\hat{h}(0) = \sqrt{2}$. Such a filter $h[n]$ and the filter $g[n]$ specified by (\ref{eqn:ac1}) constitute the scaling and wavelet filters of an MRA, respectively. The corresponding scaling and wavelet {\it functions} are obtained by iterating over the resolution parameter $j$ in the two-scale equation, which corresponds to an infinite cascade of the filterbank as illustrated in \figurename{~\ref{fig:cascade}}. More precisely, in the Fourier domain, the infinite cascade takes the form:
	\begin{flalign}
	\label{eqn:filttowave}
	\hat{\phi}(\omega) = \prod\limits_{j=1}^{\infty}\frac{\hat{h}\left(2^{-j}\omega\right)}{\sqrt{2}}\text{;} \quad
	\hat{\psi}(\omega) = \frac{1}{\sqrt{2}} \hat{g}\left(\frac{\omega}{2}\right) \hat{\phi}\left(\frac{\omega}{2}\right).
	\end{flalign}\\
	\indent Substituting $\hat{h}(0) = \sqrt{2}$ into (\ref{eqn:qmf}) yields $\hat{h}(\pi) = 0$, implying that $h[n]$ is lowpass, and has at least one root at $\omega = \pi$. When $\hat{h}(\omega)$ has $p$ roots at $\omega = \pi$, $\phi(t)$ obtained from $h[n]$ using (\ref{eqn:filttowave}) satisfies the Strang-Fix conditions \cite{strang2011fourier},  $\hat{g}(\omega)$ has $p$ roots at $\omega = 0$, and $\psi(t)$ has $p$ vanishing moments (cf. Theorem 7.4, \cite{mallat2008wavelet}), i.e.,
$$	\int\limits_{-\infty}^{\infty} t^k \psi(t)\, \mathrm{d}t = 0, \text{ for } 0 \leq k < p.$$
	The vanishing moments property empowers wavelets to annihilate polynomials of order $p-1$ or less, effectively resulting in parsimonious representations for smooth signals, i.e., signals that exhibit a high degree of regularity. In fact, this property lies at the heart of wavelet-based sparse representations for most naturally occurring signals and images.
	
	\subsection{From Biorthogonal Filterbanks to Wavelets}
	\indent A biorthogonal filterbank yields an MRA when the filters $h[n]$ and $\tilde{h}[n]$ satisfy the PR-1 condition (\ref{eqn:pr1}) as well as the following necessary conditions:
	\begin{flalign}
	    \label{eqn:necessary}
	    \hat{h}(0) = \sqrt{2}, \text{ and } \hat{\tilde{h}}(0) = \sqrt{2}.
	\end{flalign}
	The synthesis filters $\tilde{h}[n]$ and $\tilde{g}[n]$ give rise to a dual of the MRA generated by the analysis filters. The synthesis scaling and wavelet functions obtained by the infinite cascade are specified as
	\begin{flalign}
	    \label{eqn:filttowave_syn}
    	\hat{\tilde{\phi}}(\omega) = \prod\limits_{j=1}^{\infty}\frac{\hat{\tilde{h}}\left(2^{-j}\omega\right)}{\sqrt{2}}\text{,} \quad
    	\hat{\tilde{\psi}}(\omega) = \frac{1}{\sqrt{2}} \hat{\tilde{g}}\left(\frac{\omega}{2}\right) \hat{\tilde{\phi}}\left(\frac{\omega}{2}\right),
	\end{flalign}
	respectively. The number of vanishing moments of the wavelet $\tilde{\psi}(t)$ is controlled by the number of zeros of $\hat{\tilde{h}}(\omega)$ at $\omega = \pi$.\\
	\indent Unlike the orthogonal filterbank case, the infinite cascades in the biorthogonal case may not converge in $L^2(\mathbb{R})$, even though $h[n]$ and $\tilde{h}[n]$ satisfy the required conditions. Cohen and Daubechies give a sufficient condition on the filters $h[n]$ and $\tilde{h}[n]$ \cite{cohen1992stability} that allows one to check whether the infinite cascade indeed converges to a finite-energy function. The check involves constructing Lawton matrices for $h[n]$ and $\tilde{h}[n]$. The Lawton matrix
	$\boldsymbol{\Lambda}_{h} \in \mathbb{R}^{(2l - 1) \times (2l - 1)}$ for a filter $h[n]$ such that $\hat{h}(0) = \sqrt{2}$ is constructed as follows:
	\begin{flalign}
	%\nonumber
	\label{eqn:lawton}
	\Lambda_{h} = \begin{bmatrix} 
	r_{hh}[l-1] & 0 & \ldots & 0\\
	r_{hh}[l-3] & r_{hh}[l-2] & \ldots & 0\\
	\vdots & \vdots & \ddots & \vdots \\
	0 & 0 & \ldots & r_{hh}[l-1]
	\end{bmatrix},
	\end{flalign}
	where $r_{hh}$ is the autocorrelation sequence of $h[n]$; similarly for ${\tilde h}[n]$. Cohen and Daubechies showed that if the Lawton matrices corresponding to $h[n]$ and $\tilde{h}[n]$ have a non-degenerate eigenspace corresponding to the eigenvalue $1$, with all other eigenvalues being lesser than $1$, then the infinite cascade will converge to scaling and wavelet functions that form stable biorthogonal bases.\\
	\indent In summary, wavelet bases are constructed through the design of appropriately constrained PRFBs. While orthogonal wavelet bases originate from the design of a single filter $h[n]$, the biorthogonal ones require two filters, $h[n]$ and $\tilde{h}[n]$. In the following, we rely on these connections to optimize autoencoders that generate wavelets possessing the desired properties.
	
	\section{The Wavelet Learning Framework}
	\label{sec:wv_lf}
\begin{figure}[t]
		\centering
		\includegraphics[width=3.2in]{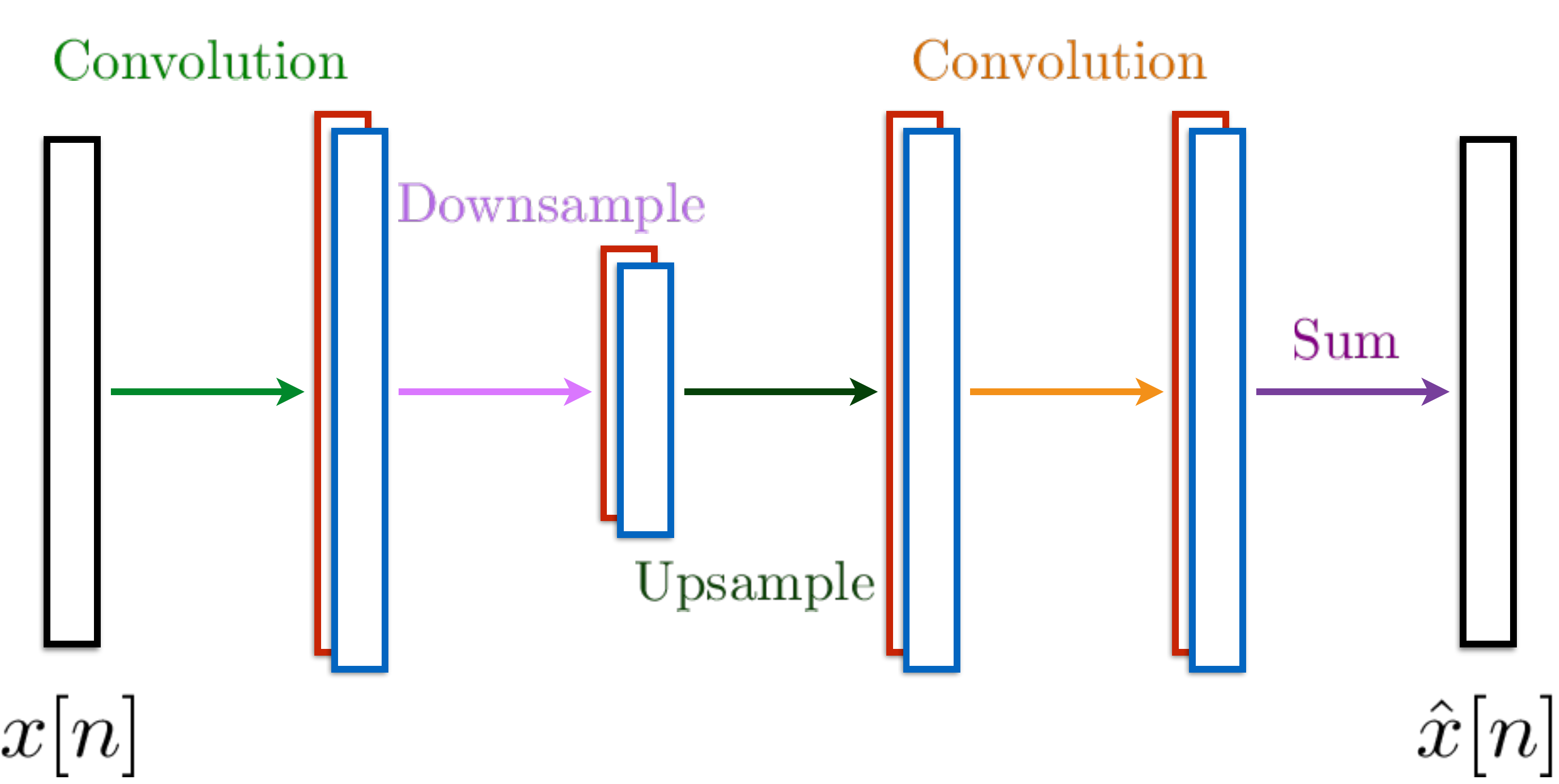}
		\caption{(Color online) A two-channel filterbank interpreted and  implemented as a convolutional autoencoder. Constraints are imposed on the filters of each convolution block so as to learn wavelets with the desired properties.}
		\label{fig:fbae}
	\end{figure}

    The starting point for the proposed wavelet learning approach is an autoencoder. In the standard scenario, an autoencoder is a nonlinear transformation achieved by means of a neural network. It can be viewed as a nonlinear generalization of principal component analysis \cite{hinton2006reducing}. Autoencoders are central objects in {\it representation learning} and are used for discovering latent features in the input data. An autoencoder comprises an encoder and a decoder. The encoder performs dimensionality reduction, which is typically accomplished using a bottleneck layer, or a sparse hidden layer \cite{ranzato2008sparse}. The decoder learns to map the compressed representation to the encoder's input. The learnt representations are useful in several practical applications --- compression \cite{theis2017lossy}, denoising \cite{vincent2010stacked}, anomaly detection \cite{ribeiro2018study}, etc.\\
    \indent We view the two-channel PRFB as a specific type of a {\it convolutional autoencoder}, henceforth referred to as the {\it filterbank autoencoder} in this paper. For the filterbank autoencoder illustrated in \figurename{~\ref{fig:fbae}}, the filtering operations are implemented as convolutional layers. The non-linear subsampling methods such as max-pooling and average-pooling typically used in deep neural networks are replaced by a linear, shift-varying downsampling operation. There is a key difference though. While an autoencoder is trained to minimize the reconstruction loss on a given dataset and rarely offers perfect reconstruction, we train a filterbank autoencoder to yield perfect reconstruction no matter what the input is, including random noise. We leverage this property to learn wavelet generating PRFBs using Gaussian vectors by imposing appropriate constraints on the filters.
    \subsection{The Filterbank Autoencoder Model}
    \label{sec:fb_ae}
	We consider finite-length filters in the filterbank autoencoder. An unconstrained filterbank autoencoder has a set of four learnable filters
	$\Theta_u = \left\{ \boldsymbol{h}, \boldsymbol{\tilde{g}} \in \mathbb{R}^{l}, \boldsymbol{\tilde{h}}, \boldsymbol{g} \in \mathbb{R}^{\tilde{l}} \right\}$
	for chosen lengths $l$ and $\tilde{l}$. One could constrain the filters to learn a {\it biorthogonal filterbank autoencoder}, in which only the filters $\boldsymbol{h}$ and $\boldsymbol{\tilde{h}}$ are learnt, whereas $\boldsymbol{g}$ and $\boldsymbol{\tilde{g}}$ are fixed according to (\ref{eqn:ac1}) and (\ref{eqn:ac2}). The filters $h[n]$ and $\tilde{g}[n]$  then have the same support size, as do the filters $\tilde{h}[n]$ and $g[n]$. The dimensionality of the search space for the filters is reduced by a factor of two in the biorthogonal setting: $	\Theta_b = \left\{ \boldsymbol{h} \in \mathbb{R}^{l}, \boldsymbol{\tilde{h}} \in \mathbb{R}^{\tilde{l}} \right\}.$
    Proceeding further, one could design an {\it orthogonal filterbank autoencoder} by setting $\boldsymbol{\tilde{h}} = \boldsymbol{h}$, with $\boldsymbol{g}$ and $\boldsymbol{\tilde{g}}$ specified by (\ref{eqn:ac1}) and (\ref{eqn:ac2}), respectively, which requires learning only a single filter $	\Theta_o = \left\{ \boldsymbol{h} \in \mathbb{R}^{l} \right\}.$
    
    \subsection{Incorporating Vanishing Moments}
	\label{sec:vm}
\begin{figure}[t]
		\centering
		\centerline{\includegraphics[width=3.2in]{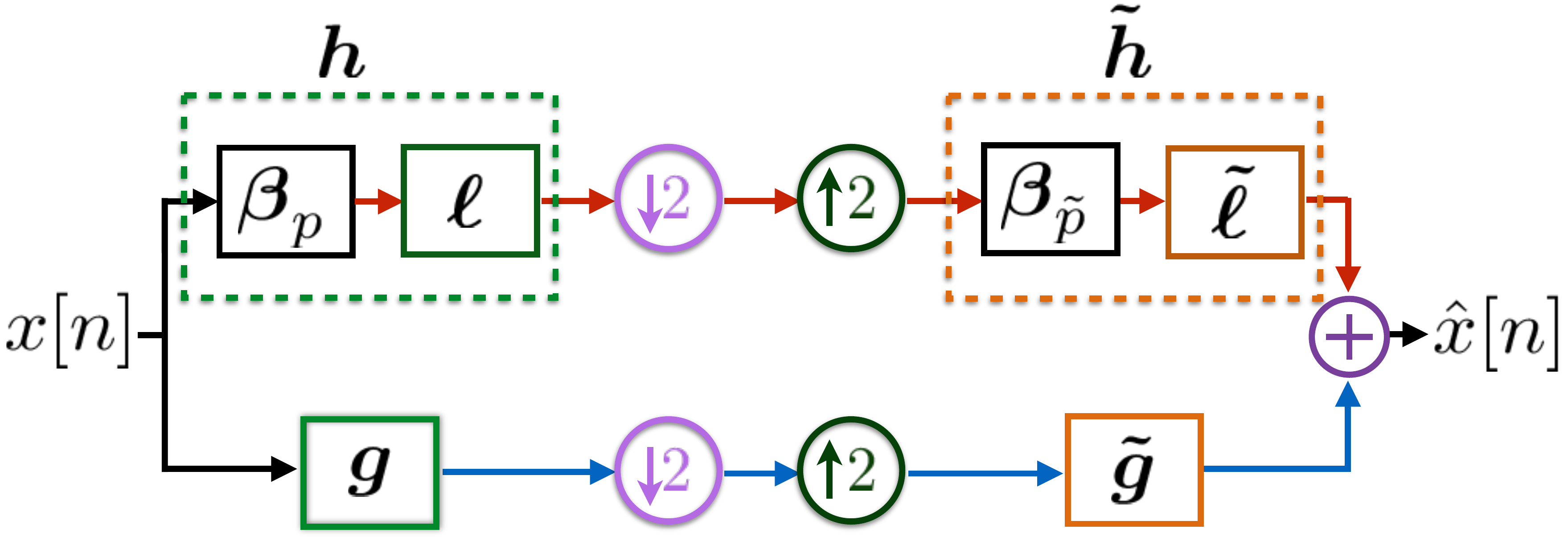}}
	    \caption{(Color online) The filterbank autoencoder with the vanishing moments constraint incorporated. The analysis filter $\boldsymbol{h}$ is expressed as a cascade of a $p^{\text{th}}$-order discrete B-spline $\boldsymbol{\beta}_{p}$ (the fixed part) and a filter $\boldsymbol{\ell}$ (the learnable part). Similarly, the synthesis filter $\boldsymbol{\tilde{h}}$ is expressed as a cascade of a $\tilde{p}^{\text{th}}$-order discrete B-spline $\boldsymbol{\tilde{\beta}}_{\tilde{p}}$ (fixed) and a learnable filter $\boldsymbol{\tilde{\ell}}$. The color scheme in Figures~\ref{fig:fbae} and \ref{fig:vm} is chosen such that the equivalent units are assigned the same color across the two figures in order to facilitate comparison and to appreciate the equivalence.}
	    \label{fig:vm}
    \end{figure}

	Not all PRFBs generate a valid multiresolution approximation. The filters $\boldsymbol{h}$ and $\boldsymbol{\tilde{h}}$ must additionally have at least one zero at $\omega = \pi$, from Equations~\eqref{eqn:qmf} and \eqref{eqn:necessary}. In the case of a wavelet with $p$ vanishing moments, the filter must have $p$ zeros at $\omega=\pi$, which leads to the following factorization of the filter $\hat {h}(\omega)$:
	\begin{equation}
	    \label{eqn:h_form_fourier}
	    \hat {h}(\omega)=\left(\frac{1+\emjw}{2}\right)^p \cdot \hat{\ell}(\omega).
	\end{equation}
	A similar factorization holds for $\hat{\tilde{h}}(\omega)$. Daubechies' construction of compactly supported orthonormal wavelets is based on this factorization. The $\hat{\ell}(\omega)$ part of the refinement filter is determined subject to the PR conditions. Unser and Blu \cite{unser2003wavelet} showed that the continuous-domain scaling function may also be expressed as the convolution of a spline component and a distributional component, and that the spline is completely responsible for several key properties of the corresponding wavelet, including vanishing moments, order of approximation, annihilation of polynomials, regularity etc. They argued that $\left(\frac{1+\emjw}{2}\right)^p$ contributes to the B-spline part and $\hat{\ell}(\omega)$ to the distributional part. For a comprehensive review on B-splines, the reader is referred to \cite{unser1993ba, unser1993bb, unser1999splines}.\\
	\indent In the learning framework that we are proposing, what this means is that the factor $\hat{\beta}_p (\omega) := \left(\frac{1+\emjw}{2}\right)^p$ is predetermined and fixed if one desires a wavelet with $p$ vanishing moments, and the {\it learnable component} is the distributional part $\hat{\ell}(\omega)$. In fact, the fixed factor is essentially a $p^\text{th}$-order discrete B-spline. Based on these considerations, we express the filters as follows:
	\begin{flalign}
	\label{eqn:h_form}
	\boldsymbol{h} = \boldsymbol{\beta}_p * \boldsymbol{\ell}, \quad \text{and} \quad  	\boldsymbol{\tilde{h}} = \boldsymbol{\beta}_{\tilde{p}} * \boldsymbol{\tilde{\ell}},
	\end{flalign}
	where $*$ denotes the 1-D convolution operation, the vector $\boldsymbol{\beta}_p \in \mathbb{R}^{p + 1}$ is the discrete $p^{\text{th}}$-order B-spline, and the vectors $\boldsymbol{\ell} \in \mathbb{R}^{l - p}$ and $\boldsymbol{\tilde{\ell}} \in \mathbb{R}^{\tilde{l} - \tilde{p}}$ are learnt. The filterbank autoencoder that incorporates the {\it vanishing moments constraint} is shown in \figurename{~\ref{fig:vm}}. The parameters $p$ and $\tilde{p}$ are the number of vanishing moments of the wavelets $\psi(t)$ and $\tilde{\psi}(t)$, respectively, generated by these filters. In summary, the parameter sets for learning orthogonal and biorthogonal autoencoders with vanishing moments are 
	\begin{flalign}
	\label{eqn:vc}
	\nonumber
	\Theta_o^{p} &= \{ \boldsymbol{h} \in \mathbb{R}^{l} \mid \boldsymbol{h} = \boldsymbol{\beta}_p * \boldsymbol{\ell}\} \text{ and }\\ \Theta_b^{p, \tilde{p}} &= \{ \boldsymbol{h} \in \mathbb{R}^{l}, \boldsymbol{\tilde{h}} \in \mathbb{R}^{\tilde{l}} \mid \boldsymbol{h} = \boldsymbol{\beta}_p * \boldsymbol{\ell}, \boldsymbol{\tilde{h}} = \boldsymbol{\beta}_{\tilde{p}} * \boldsymbol{\tilde{\ell}} \},
	\end{flalign}
	respectively.
	
	\subsection{The Optimization Problem}
	We use the mean-squared error loss for training the filterbank autoencoder. Given data $\mathcal{X} = \{ \boldsymbol{x}_j \in \mathbb{R}^{s} \}_{j = 1}^{m}$, the loss is defined as
	\begin{align}
	    \label{eqn:loss}
    	\mathcal{L}(\mathcal{X}; \Theta) = \frac{1}{m}\sum_{j = 1}^{m} \|\boldsymbol{x}_j - \boldsymbol{\hat{x}}_j(\Theta)\|_2^2,
	\end{align}
	where $\Theta$ denotes the parameter set to optimize (the filter coefficients) and $\boldsymbol{\hat{x}}_j(\Theta)$ is the output of the autoencoder corresponding to the input $\boldsymbol{x}_j$. In the orthogonal wavelet case with $p$ vanishing moments, $\Theta = \Theta_o^{p}$, and in the biorthogonal case with $(p,\tilde{p})$ vanishing moments, $\Theta = \Theta_b^{p, \tilde{p}}$. The objective is to minimize
	$\mathcal{L}(\mathcal{X}; \Theta)$ with respect to $\Theta$.
	The Adam optimization algorithm \cite{kingma2014adam} is used for carrying out the minimization. Adam adapts the learning rate for each model parameter to expedite  convergence.\\
	\indent Since the filterbank autoencoder is linear, one could write $\boldsymbol{x}_j - \boldsymbol{\hat{x}}_j(\Theta) = \boldsymbol{B}_{\Theta} \boldsymbol{x}_j$. When $\boldsymbol{B}_{\Theta}$ equals $\boldsymbol{0}_s$, the filters in $\Theta$ form a perfect reconstruction filterbank. This result is captured in the form of the following proposition. The quantity $\boldsymbol{B}_{\Theta}$ is determined by the filters in the parameter set $\Theta$. For the sake of brevity of notation, in the following analysis, we drop the subscript $\Theta$ in $\boldsymbol{B}_{\Theta}$.
	\begin{prop}
	    \label{prop:loss_form}
	    For a set of filters $\Theta= \left\{ \boldsymbol{h}, \boldsymbol{\tilde{g}} \in \mathbb{R}^{l}, \boldsymbol{\tilde{h}}, \boldsymbol{g} \in \mathbb{R}^{\tilde{l}} \right\}$, and a dataset $\mathcal{X}= \{ \boldsymbol{x}_j \in \mathbb{R}^{s} \}_{j = 1}^{m}$, there exists a matrix $\boldsymbol{B} \in \mathbb{R}^{s \times s}$ such that the loss function defined in \eqref{eqn:loss} can be expressed as
	    \begin{align}
	        \label{eqn:loss_form}
	        \mathcal{L}(\mathcal{X}; \Theta) = \frac{1}{m}\sum_{j = 1}^{m} \|\boldsymbol{B} \boldsymbol{x}_j\|_2^2.
	    \end{align}
	    If the filters form a PRFB, then $\boldsymbol{B} = \boldsymbol{0}_s$ and vice versa.
	\end{prop}
	\begin{IEEEproof}
        The proof is given in Appendix A.%\ref{sec:appendix_form}.
	\end{IEEEproof}
	\indent When the filters form a PRFB, the loss  $\mathcal{L}(\mathcal{X}; \Theta)$ becomes zero and vice versa. Hence, the loss as defined in \eqref{eqn:loss} is indeed appropriate to learn a PRFB. The PRFB property must hold regardless of the data used for training the filterbank autoencoder. This brings us to the question about the choice of the data. We show that choosing random Gaussian vectors suffices for the purpose of training. The consequent statistical guarantees are reassuring.
	
	\subsection{Training Data}
	In our experiments, we choose a dataset containing $m = s$ Gaussian vectors,  i.e., $ \mathcal{X} = \{\boldsymbol{x}_j \in \mathbb{R}^s \mid \boldsymbol{x}_j \sim \mathcal{N}(\boldsymbol{0}, \boldsymbol{I}_s)\}_{j = 1}^{s}, $ where each entry $x_i$ of $\boldsymbol{x} \in \mathcal{X}$ is sampled independently and identically from a zero-mean, unit-variance Gaussian distribution. A training dataset comprising Gaussian vectors has two benefits: first, $s$ randomly sampled Gaussian vectors are nearly orthogonal with high probability; and second, the squared-length of linearly transformed Gaussian vectors, i.e., $\|\boldsymbol{B x}\|_2^2$ concentrates around its mean value with high probability.\\
	\indent The expected squared-length of $\boldsymbol{x} \in \mathcal{X}$ is given by $\mathbb{E}[\|\boldsymbol{x}\|_2^2] = \mathbb{E}\left[\sum_{i = 1}^{s} x_i^2\right] = s$, using the independence and unit variance properties of $x_i$. We recall the {\it Gaussian annulus theorem} (cf. Theorem 2.9, \cite{blum2020foundations}) below.
    \begin{theorem} \cite{blum2020foundations}
        \label{thm:annulus}
        For an s-dimensional spherical Gaussian centered on the origin and with unit variance in each direction, for any $k \leq \sqrt{s}$, all but at most $3 e^{-\frac{ k^2}{96}}$ of the probability mass lies within the annulus $\sqrt{s} - k \leq \|\boldsymbol{x}\|_2 \leq \sqrt{s} + k$.
    \end{theorem}
    The theorem states that, with high probability, the length of the Gaussian vector $\boldsymbol{x}$ is $\sqrt{s}$. The expected squared Euclidean distance between two independent Gaussian vectors with zero-mean and unit-variance entries $\boldsymbol{x}$ and $\boldsymbol{y}$ is given by
    \begin{align}
    \mathbb{E}[\|\boldsymbol{x-y}\|_2^2] &= \mathbb{E}\left[\displaystyle\sum_{i = 1}^{s} (x_i - y_i)^2\right] \nonumber \\
    &= \displaystyle\sum_{i = 1}^{s} \left( \mathbb{E}\left[x_i^2\right] + \mathbb{E}\left[y_i^2\right] - 2 \mathbb{E}[x_i] \mathbb{E}[y_i] \right) =  2s.\nonumber
    \end{align}
    Since the squared-length of each vector $\boldsymbol{x}$ and $\boldsymbol{y}$ is close to $s$ with high probability, and the expected distance between them is close to $2s$, using Pythagoras theorem, we can conclude that the vectors are orthogonal with a high probability. We recollect another important result (Theorem 2.8 from \cite{blum2020foundations}), which proves that $s$ vectors drawn at random from the unit ball are orthogonal with high probability. Owing to spherical symmetry of the standard Gaussian distribution, $\mathcal{N}(\boldsymbol{0}, \boldsymbol{I}_s)$, vectors on the unit ball in $s$-dimensions can be obtained by drawing $s$-dimensional Gaussian vectors and rescaling them to possess unit norm. 
    \begin{theorem} \cite{blum2020foundations}
        Consider drawing $m$ points $\boldsymbol{x}_1, \ldots, \boldsymbol{x}_m \in \mathbb{R}^s$ at random from the unit ball. With probability $1 - \mathcal{O}(\frac{1}{m})$,
        \begin{align*}
          \|\boldsymbol{x}_j\|_2 \geq 1 - \frac{2 \ln m}{s}, &\text{ for all } j, \text{ and } \\
          |\langle \boldsymbol{x}_i, \boldsymbol{x}_j \rangle| \leq \frac{\sqrt{6 \ln m}}{\sqrt{s - 1}}, &\text{ for all } i \neq j.
        \end{align*}
    \end{theorem}
   \indent Based on the orthogonality property of the vectors in $\mathcal{X}$, the loss function in \eqref{eqn:loss_form} could be interpreted as the square of the Hilbert-Schmidt norm of the finite-dimensional operator $\boldsymbol{B}$. We show that $\mathcal{L}(\mathcal{X}; \Theta)$ is concentrated about its expected value with high probability. The expectation of the loss is $\| \boldsymbol{B} \|_{\mathrm{F}}^2$, where $\|\cdot\|_\textsc{F}$ denotes the Frobenius norm, as justified below:
   	\begin{align*}
	    \mathbb{E}\{\mathcal{L}(\mathcal{X}; \Theta)\} &= \mathbb{E}\left\{\frac{1}{s}\sum_{j = 1}^{s} \|\boldsymbol{B} \boldsymbol{x}_j\|_2^2\right\}, \\
	    &= \mathbb{E}\{\|\boldsymbol{B} \boldsymbol{x}\|_2^2\} = \mathbb{E}\{\boldsymbol{x}^\textsc{T}\boldsymbol{B}^\textsc{T}\boldsymbol{B}\boldsymbol{x}\}\\
        %&= \sum_{i,j} (\boldsymbol{B}^\textsc{T}\boldsymbol{B})_{i,j} \mathbb{E}\{{x}_i{x}_j\}\\
	    &= \sum_i (\boldsymbol{B}^\textsc{T}\boldsymbol{B})_{i,i} = \|\boldsymbol{B}\|_{\textsc{F}}^2,
\end{align*}
    where we have used the property that the entries of $\boldsymbol{x}$ are i.i.d. 
    The random vector $\boldsymbol{B} \boldsymbol{x}$ is Gaussian, with mean $\boldsymbol{0}$ and covariance matrix $\boldsymbol{B} \boldsymbol{B}^{\mathrm{T}}$. Only when $\boldsymbol{B}\boldsymbol{B}^{\mathrm{T}}$ is a diagonal matrix will the entries of $\boldsymbol{B} \boldsymbol{x}$ have unit variance and be independent. In the special case where $\boldsymbol{B}$ is a unitary matrix, $\boldsymbol{B} \boldsymbol{x}$ follows the standard Gaussian distribution, more precisely, $\boldsymbol{B} \boldsymbol{x} \sim \mathcal{N}(\boldsymbol{0}, \boldsymbol{I}_s)$, which is due to the \emph{rotational invariance} of the standard Gaussian. The exact nature of the concentration of the loss about its mean is specified next.
    \begin{theorem}
        \label{thm:conc_loss}
        For the loss function $\mathcal{L}$ defined in \eqref{eqn:loss_form}, where $\mathcal{X}$ comprises standard Gaussian vectors, the deviation of the loss from its expected value is bounded in probability as follows:
        \begin{align*}
        \mathbb{P}(\big| \mathcal{L} - \mathbb{E} \{ \mathcal{L} \} \big| \geq k) \leq \begin{cases} 
                2 e^{\frac{-k^2 s}{8 \|\boldsymbol{B}\|_{2}^{2} \|\boldsymbol{B}\|_{\textsc{F}}^{2}}}, \, &k \leq \|\boldsymbol{B}\|_{\textsc{F}}^2, \\
                2 e^{\frac{-k s}{8 \|\boldsymbol{B}\|_{2}^{2}}}, \, &k > \|\boldsymbol{B}\|_{\textsc{F}}^2.
            \end{cases}
    \end{align*}
    \end{theorem}
    \begin{IEEEproof}
            The proof is given in Appendix B.%\ref{sec:appendix_conc}. 
    \end{IEEEproof}
    
    A more general version of Theorem \ref{thm:conc_loss} is the Hanson-Wright inequality \cite{rudelson2013hanson}, which extends the argument to linear transformations of sub-Gaussian vectors. For small deviations about the mean ($k \leq \|\boldsymbol{B}\|_{\textsc{F}}^2$), the concentration behavior of $\|\boldsymbol{B} \boldsymbol{x}\|_2^2$ is similar to that of a Gaussian, while for large deviations, the tail probability is heavier than that for a Gaussian. More specifically, $\|\boldsymbol{B} \boldsymbol{x}\|_2^2$ is a \emph{sub-exponential} random variable (cf. Theorem 2.13, \cite{wainwright2019high}).
    The preceding theorem states that, for standard Gaussian data $\mathcal{X}$ and $k \leq \|\boldsymbol{B}\|_{\textsc{F}}^2$, all but at most $2 e^{\frac{-k^2 s}{8 \|\boldsymbol{B}\|_{2}^{2} \|\boldsymbol{B}\|_{\text{F}}^{2}}}$ of the probability mass of the loss lies within the annulus $\|\boldsymbol{B}\|_{\textsc{F}}^2 - k \leq \mathcal{L}(\mathcal{X}; \Theta) \leq \|\boldsymbol{B}\|_{\textsc{F}}^2 + k$. Proposition \ref{prop:loss_form} indicates that $\|\boldsymbol{B}\|_{\textsc{F}}^2$ is a direct measure of how close the filters are to achieving perfect reconstruction. The high concentration of the loss about its mean for Gaussian data goes to show that minimizing an instance of the loss is, with high probability, equivalent to minimizing its expectation, which is an assurance that the filters are being steered toward achieving PR.
    
   \begin{figure*}[t]
		\centering	\includegraphics[width=6.4in]{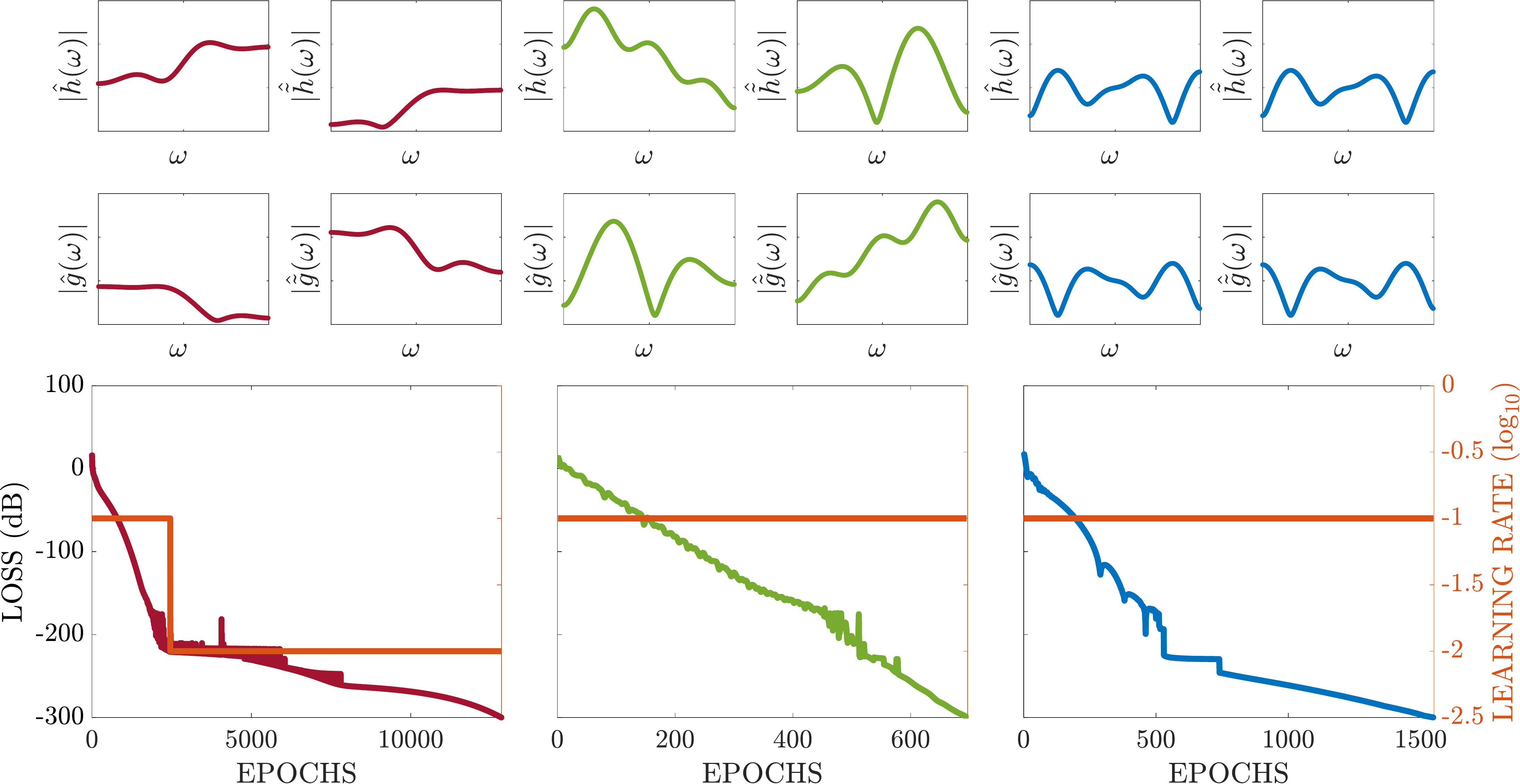}
		\caption{(Color online) Learning filterbank autoencoders: The results of learning unconstrained (red), biorthogonal constrained (green), and orthogonal constrained (blue) filterbank autoencoders are shown. The frequency responses of the learnt filters are displayed in the top two rows ($\omega \in [0, \pi]$; the $y$-axis limits are $[0, 3]$). The loss as a function of the epochs and the learning rate (right-hand side $y$-axis) for Adam on a log-scale are shown. Imposing biorthogonality/orthogonality constraints leads to faster convergence compared with the unconstrained case. Note that the $x$-axis limits in the loss vs. epochs plots are different in the three cases.}
		\label{fig:fb_learn}
	\end{figure*}
   \subsection{Learning Rate Schedule}
	While using optimizers such as Adam \cite{kingma2014adam}, a monotonically decreasing loss is not guaranteed, and tuning the learning rate parameter $\eta$ during training becomes necessary. A high learning rate may accelerate the optimization in the initial phase, but would ultimately lead to a large deviation away from the minimum loss value. The back-tracking line-search method (cf. Section 9.2 of \cite{boyd2004convex}) finds an appropriate learning rate $\eta^{(k)}$ at iteration $k$ by starting with $\eta^{(k)} = 1$ and iteratively dividing $\eta^{(k)}$ by a factor $\alpha > 1$ as long as the gradient-descent step does not result in a reduction of the loss. However, this method requires one to compute the output of the autoencoder several times at each iteration $k$ before the next update is computed. We use a heuristic that requires computing the forward pass only once at iteration $k$, by tolerating the deviation from the minimum loss value from iterations $1$ to $k$, up to a factor $\tau$.\\
	\indent Let the loss at iteration $k$ be denoted by $\mathcal{L}^{(k)}(\mathcal{X}; \Theta)$. We keep track of the minimum value of the loss as well as the corresponding model state. Let the minimum value of the loss function encountered up to the current iteration be $\mathcal{L}^{(k_{\text{min}})}(\mathcal{X}; \Theta)$, occurring at iteration index $k_{\text{min}}$. At iteration $k > k_{\text{min}}$, if $\log_{10}\left( \frac{\mathcal{L}^{(k)}(\mathcal{X}; \Theta)}{\mathcal{L}^{(k_{\text{min}})}(\mathcal{X}; \Theta)} \right) > \tau$, $\eta$ is reduced by a certain factor $\alpha$, i.e., $\eta \leftarrow \frac{\eta}{\alpha}$, and the training is resumed after resetting the model state to that of iteration $k_{\text{min}}$. This procedure effectively reduces the learning-rate every time a large deviation from the minimum occurs, and the reduced learning-rate is used until either another deviation occurs, or until the stopping criterion is satisfied. In our experiments, we used a threshold of $\tau = 6$ and a factor $\alpha = 10$.
	
	\subsection{Stopping Criteria}
	We use two stopping criteria and terminate the training when either of them is met. The first criterion measures the relative change in the loss value for 100 consecutive iterations. The training is stopped at iteration $k$ if the percentage change in the loss value falls below a certain threshold $\delta$ for $100$ consecutive iterations preceding $k$, i.e., if for all $i$ such that $ 0 \leq i \leq 99$, the following happens:
	$$
	    \frac{|\mathcal{L}^{(k-i)}(\mathcal{X}; \Theta) - \mathcal{L}^{(k-i-1)}(\mathcal{X}; \Theta)|}{\mathcal{L}^{(k-i-1)}(\mathcal{X}; \Theta)} < \delta.
	$$
	\indent The second criterion is an absolute one and checks for the loss falling below a preset threshold $\epsilon$, i.e., $\mathcal{L}^{(k)}(\mathcal{X}; \Theta) < \epsilon$. 	In our experiments, we set $\delta = 10^{-5}$ and $\epsilon = 10^{-15}$.
	
	\subsection{Performance Measures}
	\label{sec:perf_measures}
	We employ two performance measures to determine how close to PR the learnt filterbank autoencoders are. The first one is the signal-to-reconstruction error ratio (SRER) defined as $$\text{SRER} = 20 \log_{10}\left( \frac{1}{s} \sum\limits_{i = 1}^{s}  \frac{\|\boldsymbol{x}_i\|_2}{\|\boldsymbol{x}_i - \tilde{\boldsymbol{x}}_i\|_2}\right)\,\text{dB}.$$
	A test set of white Gaussian vectors $\left\{\boldsymbol{x}_i \sim \mathcal{N}(\boldsymbol{0}_{1000}, \boldsymbol{I}_{1000})\right\}_{1 \leq i \leq 1000}$ is used to compute the SRER. An SRER greater than $144.50$ dB for single-precision floats and an SRER greater than $319.09$ dB for double-precision floats indicates machine epsilon, or maximum relative error due to rounding, according to the IEEE 754 standard \cite{ieee754}.\\
	\indent An ideal filterbank autoencoder must satisfy the PR-1 condition for all $\omega \in [0, 2\pi]$. The second performance measure checks the PR-1 condition considering samples of $\omega \in [0, 2\pi]$:
	$$
	\Delta_{\text{pr}} = \frac{1}{N}\sum\limits_{k = 0}^{N-1} \left(\hat{h}^*(\omega_k) \hat{\tilde{h}}(\omega_k) + \hat{h}^*(\omega_k + \pi) \hat{\tilde{h}}(\omega_k + \pi) - 2\right)^2,
	$$
    where $h$ and $\tilde{h}$ are the learnt analysis and synthesis lowpass filters, respectively, and $\omega_k = \frac{2 \pi k}{N}, 0 \leq k \leq N-1, N = l + \tilde{l} - 1 $.

	\begin{figure*}[t]
		\centering
		\includegraphics[width=5in]{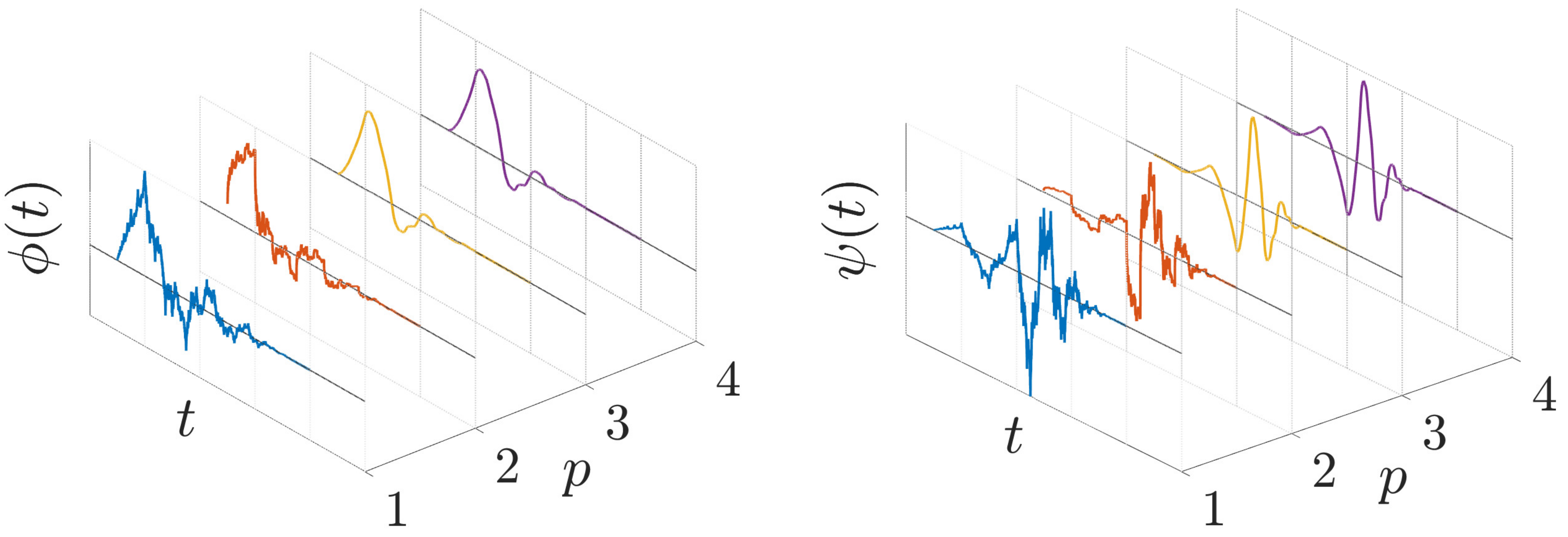}
		\caption{The learnt orthogonal scaling (left) and wavelet (right) functions, with filter length $l = 8$ and vanishing moments $p$ varying from $1$ to $4$. Observe that the smoothness of both the scaling and wavelet functions increases as the number of vanishing moments imposed increases.}
		\label{fig:vm_increase}
	\end{figure*}
	
	\subsection{Experimental Validation}
	We present the results of learning filterbank autoencoders corresponding to the three flavors: unconstrained (parameter set $\Theta_u$), biorthogonal ($\Theta_b$), and orthogonal ($\Theta_o$). The models were implemented using the Tensorflow Python library \cite{tensorflow2015-whitepaper} and run on a computer with an i7 processor and 8 GB RAM. The stopping criterion threshold $\epsilon$ was set to $10^{-30}$. For all three filterbanks, the SRER was greater than $200$ dB, and $\Delta_{\text{pr}}$ was of the order of $10^{-18}$.\\
	\indent The frequency responses of the filters, the training loss, and learning rate as a function of epochs are shown in \figurename{~\ref{fig:fb_learn}}. The unconstrained and biorthogonal filterbanks were initialized with the same filters $\boldsymbol{h}$ and $\boldsymbol{\tilde{h}}$, having length $8$ each, with their entries drawn from the standard normal distribution. The filter $\boldsymbol{h}$ of the orthogonal autoencoder was also initialized with the same values as the filter $\boldsymbol{h}$ for the other cases. Additionally, the filters $\boldsymbol{g}$ and $\boldsymbol{\tilde{g}}$ were initialized randomly for the unconstrained variant, while they were computed using the biorthogonal relations (\ref{eqn:ac1}) and (\ref{eqn:ac2}) for the other two models. The same dataset of $128$ random vectors was used to train all the models.
	
	We observe that each model converges to a PRFB, as indicated by the loss function going below the threshold of $\epsilon = 10^{-30}$. For the unconstrained case, the filters $\boldsymbol{g}$ and $\boldsymbol{\tilde{g}}$ were found to obey the biorthogonal relations (\ref{eqn:ac1}) and (\ref{eqn:ac2}) automatically although they were not specifically constrained to do so, indicating that training under the MSE loss and our choice of the dataset indeed enforces the perfect reconstruction property. The unconstrained model takes  longer to train than the constrained versions. Although the three models were given the same initialization, they converged to different solutions, demonstrating the non-uniqueness of the solution space. Also, the filters learnt do not have any zeros, neither at $\omega = 0$ nor at $\omega = \pi$, indicating that the learnt filterbank autoencoders, although satisfying the PR property, do not generate a {\it bona fide} multiresolution approximation.
	
	\section{From Filterbank Autoencoders to Wavelets}
	\label{sec:1d_wv}
	In this section, we explain how wavelet properties such as vanishing moments, orthogonality, symmetry, and biorthogonality could be incorporated into the filterbank autoencoder learning process. The learning paradigm, choice of the loss function and dataset are as discussed in the previous section. Once a filterbank autoencoder with the desired properties is learnt, the corresponding scaling and wavelet functions are obtained using the infinite cascade relation specified in \eqref{eqn:filttowave}. The wavelet learnt by an autoencoder for a particular setting of vanishing moments $p$ and $\tilde{p}$ need not be unique, and additional constraints may have to be imposed to arrive at wavelets with the desired properties. These nuances are discussed in the following, in which we consider learning both orthogonal (corresponding to $\Theta_o^{p}$) and biorthogonal wavelet bases (corresponding to $\Theta_b^{p, \tilde{p}}$).
	
	\subsection{Orthogonal Wavelets with $p$ Vanishing Moments}
	Orthogonal wavelets are learnt using the autoencoder model $\Theta_o^{p}$ defined in \eqref{eqn:vc}. The parameters to be specified are the length of the filter ($l$) and the number of vanishing moments ($p$). The length of $\boldsymbol{h}$ must be greater than twice the number of vanishing moments $p$ \cite{daubechies1988orthonormal}, and for a given choice of $l \geq 2 p$ and $p \geq 1$, one of several possible wavelet generating filterbanks may be learnt. The number of vanishing moments of a wavelet is related to its smoothness, as established by Tchamitchian (Theorem 7.6, \cite{mallat2008wavelet}). We demonstrate this relationship in \figurename{~\ref{fig:vm_increase}} by learning length-$8$ filters $\boldsymbol{h}$, with $p$ varying from $1$ to $4$. As expected, the smoothness of the learnt wavelet and scaling functions increases with an increase in the number of vanishing moments imposed. For $p = 4$,  the learnt wavelet closely resembles the db$4$ wavelet of the Daubechies family. This is the maximally smooth wavelet for $l = 8$ and $p = 4$. In fact, Daubechies proved that the maximally smooth wavelet for a fixed, even length $l$ of $\boldsymbol{h}$ has $p = \frac{l}{2}$ vanishing moments \cite{daubechies1988orthonormal}. We next discuss how to learn members of the Daubechies family.

	\begin{figure*}[t]
		\centering	
		\includegraphics[width=6in]{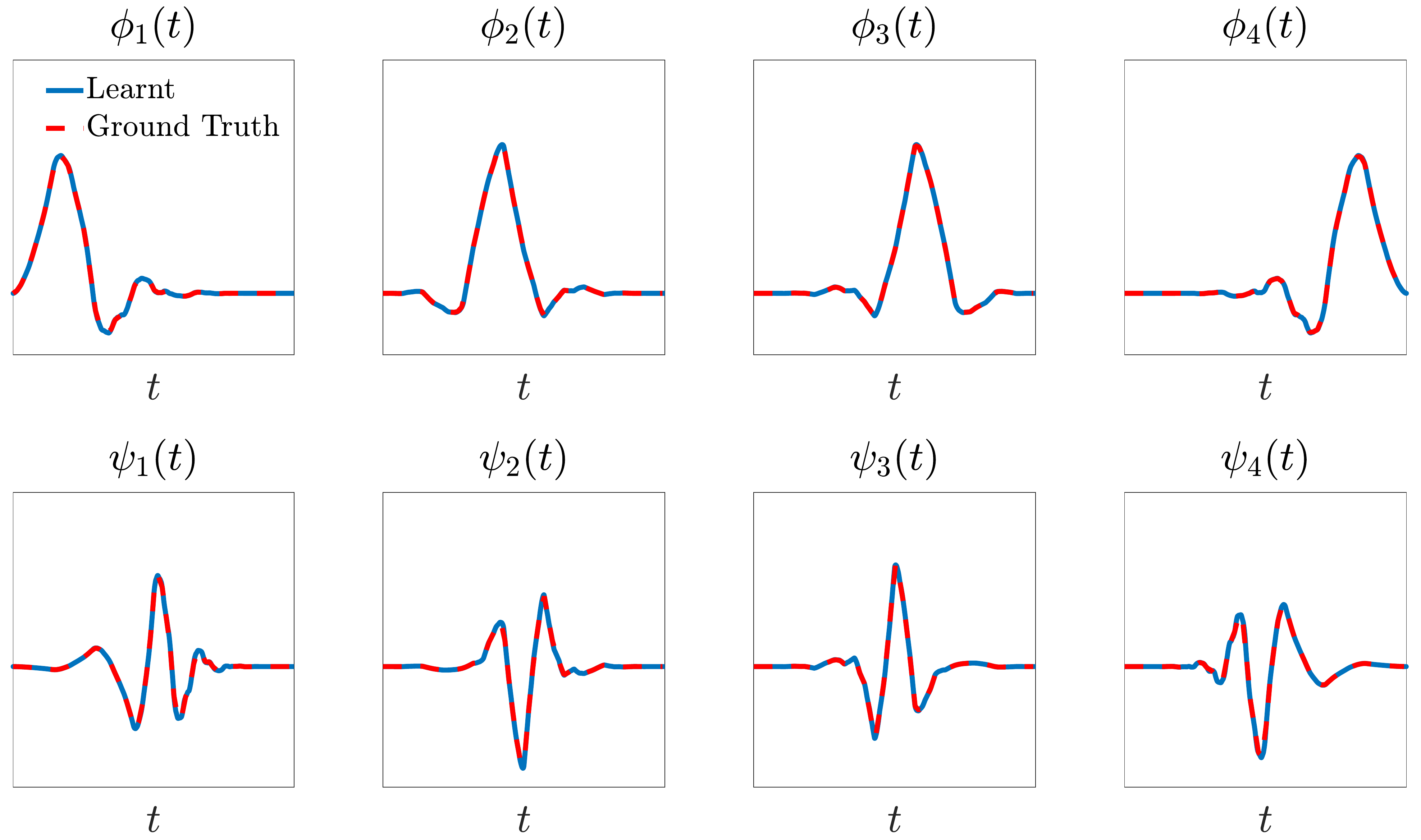}
		\caption{(Color online) Scaling (top row) and wavelet (bottom row) function pairs $(\phi_i(t), \psi_i(t)), i = 1, 2, \ldots, 4$, obtained by learning an orthogonal filterbank autoencoder (filter length $l = 8$, vanishing moments $p = 4$). Each column shows one pair obtained by a different training dataset and a different initialization of $\boldsymbol{h}$. The $x$-axis limits for all plots are $[0, 7]$, the $y$-axis limits for $\phi_i(t)$ and $\psi_i(t)$ are $[-0.5, 1.9]$ and $[-1.8, 2.6]$, respectively. As anticipated, the solution is not unique. We found that the learnt solutions obtained correspond to one out of the four possible factorizations of the polynomial $Q(y)$ in Equation~\eqref{eqn:q_y}. For reference, the scaling and wavelet functions obtained by factorizing $Q(y)$, which serve as the ground truth, are also shown. The learnt scaling and wavelet functions are nearly identical to their ground-truth counterparts.}
		\label{fig:db4_repeat}
	\end{figure*}
	
	\subsection{Daubechies Family of Orthogonal Wavelets}
	The Daubechies wavelet with $p$ vanishing moments, denoted as the db$p$ wavelet, is generated by the minimum-phase filter of length $l = 2p$ satisfying the CMF condition \eqref{eqn:qmf}. The minimum-phase criterion makes the filter real-valued and unique among the $2^{p - 1}$ possible length-$2p$ filters, complex-valued in general, having $p$ vanishing moments satisfying (\ref{eqn:qmf}) (cf. Section 4 of \cite{daubechies1988orthonormal}). We show that a filterbank autoencoder with $l = 8$ and $p = 4$ converges to one of four real-valued wavelet generating filterbanks. Over $10$ repetitions of the experiment, with different random initializations of $\boldsymbol{h}$ and different datasets, the four wavelets and scaling functions learnt are shown in \figurename{~\ref{fig:db4_repeat}}. The wavelet $\psi_1(t)$ corresponds to the minimum-phase (Daubechies) wavelet and $\psi_2(t)$ corresponds to the length-$8$ Symmlet. We also observe from \figurename{~\ref{fig:db4_repeat}} that $\phi_3(t) = \phi_2(-t)$, $\phi_4(t) = \phi_1(-t)$, $\psi_3(t) = -\psi_2(-t)$ and $\psi_4(t) = -\psi_1(-t)$. Understanding this finding further requires one to go into the details of Daubechies' construction.
	
	Considering the factorization of $\boldsymbol{h}$ specified in (\ref{eqn:h_form_fourier}), we may rewrite (\ref{eqn:pr1}) as follows:
	\begin{flalign}
	    \label{eqn:daub_form}
	    P(y)Q(1 - y) + P(1 - y)Q(y) = 2,
	\end{flalign} 
	where $y = \frac{(2 + e^{\mathrm{j} \omega} + e^{-\mathrm{j} \omega})}{4}$, $P(y) = 4^p y^p$ and $Q(y) = |\hat{\ell}(\omega)|^2$. The polynomial $Q(y)$ that satisfies \eqref{eqn:daub_form} is specified by Bezout's theorem (cf. Theorem 7.8, \cite{mallat2008wavelet}).
	\begin{theorem} \label{thm:bezout} \cite{mallat2008wavelet} Let $P_1(y)$ and $P_2(y)$ be two polynomials of degrees $p_1$ and $p_2$, respectively, and having no common zeros. Then, there exist two unique polynomials $Q_1(y)$ and $Q_2(y)$ of degrees $p_2 - 1$ and $p_1 - 1$, respectively, such that
	    \begin{align*}
	        P_1(y) Q_1(y) + P_2(y) Q_2(y) = 1.
	    \end{align*}
	\end{theorem}
	Since $P(y)$ is an order-$p$ polynomial in $y$, having no common zeros with the order-$p$ polynomial, $P(1 - y)$, $Q(y)$, and $Q(1 - y)$ are order-$(p - 1)$ polynomials. The roots of $Q(y)$, once found, must be factorized to determine $\hat{\ell}(\omega)$. However, the factorization is not unique. The trigonometric polynomial $\hat{\ell}(\omega)$ is expressed in terms of its roots, $\{ r_k \}_{k = 1}^{p - 1}$, as follows:
	\begin{flalign}
	\hat{\ell}(\omega) = \ell[0] \prod\limits_{k = 1}^{p - 1} (1 - r_k e^{-\mathrm{j} \omega}).
	\end{flalign}
	Since the complex conjugate $\hat{\ell}^*(\omega)$ has roots $\{ \frac{1}{r_k} \}_{k = 1}^{p - 1}$, $Q(y)$ is written in terms of its roots as follows:
	\begin{flalign}
	\label{eqn:q_y}
	Q(y) = |\hat{\ell}(\omega)|^2 = \left(\ell[0]\right)^2 \prod\limits_{k = 1}^{m} \left\{(1  + r_k)^2 - 4 r_k y\right\}.
	\end{flalign}
	Observe that every root $r_k$ of $\hat{\ell}(\omega)$ corresponds to the root $\frac{(1 + r_k)^2}{4 r_k}$ of the polynomial $Q(y)$. Given a root for $Q(y)$, one could obtain $r_k$ by solving a quadratic equation. One could check that both $r_k$ and $\frac{1}{r_k}$ construct the same root of $Q(y)$, implying that if we choose $r_k$ as a root of $\hat{\ell}(\omega)$, then $\frac{1}{r_k}$ is automatically a root of $\hat{\ell}^*(\omega)$. In the construction of the db$p$ wavelet, the roots $r_k$ assigned to $\hat{\ell}(\omega)$ are chosen so that $|r_k| < 1$, making $h[n]$ a minimum-phase filter.\\
	\indent For the case $p = 4$, both $Q(y)$ and $\hat{\ell}(\omega)$ have three roots, two of which are complex. Since we only consider real filters $\ell[n]$, there are four possible factorizations of $Q(y)$, which yield the four solutions learnt. Let the roots of $\hat{\ell}_1(\omega)$, which is the learnable part of the filter $\hat{h}_1(\omega)$ that generates the scaling and wavelet functions $\phi_1(t)$ and $\psi_1(t)$, respectively, be $(r_1, r_2, r_3)$, where $r_1$ is real and $r_2$ and $r_3$ form a complex-conjugate pair. We found experimentally that the roots of the learnable filters $\hat{\ell}_i(\omega)$ corresponding to $\phi_i(t)$ and $\psi_i(t), i = 2, 3, 4$, are $(\frac{1}{r_1}, r_2, r_3), (r_1, \frac{1}{r_2}, \frac{1}{r_3})$, and $(\frac{1}{r_1}, \frac{1}{r_2}, \frac{1}{r_3})$, respectively. The relationship between the roots of $\hat{\ell}_i(\omega)$ explains the relationship between the learnt scaling and wavelet functions.\\
	\indent In general, setting $m = 2p$ and learning wavelets would yield one of the possible real factorizations of the filter $\hat{\ell}(\omega)$ from the roots of $Q(y)$. Hence, in order to obtain a minimum-phase db$p$ wavelet, we train an orthogonal filterbank autoencoder with $l = 2p$ and factorize the learnt filter $\boldsymbol{h}$ to obtain its roots $\{ r_k \}_{k = 1}^{p - 1}$. The db$p$ wavelet filter is then constructed by replacing each root $r_k$ having magnitude greater than unity with $\displaystyle\frac{1}{r_k}$.
	
	\subsection{Symmetric Orthogonal Wavelets?}
	\label{subsec:symm_ortho}
	
	\begin{figure*}[t]
        \centering
        \includegraphics[width=6.5in]{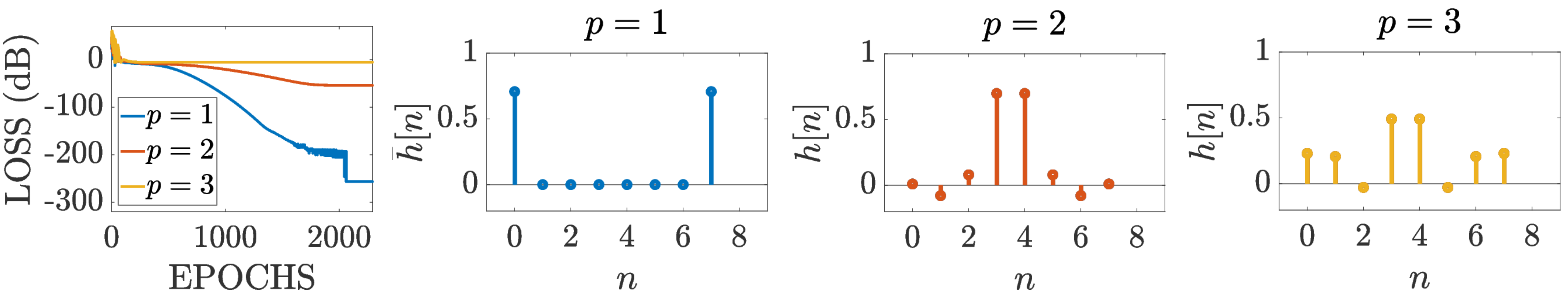}
        
        \caption{(Color online) Results pertaining to learning symmetric orthogonal wavelets for length-$8$ filters with different number of vanishing moments ($p = 1, 2, 3$). In all three cases, the learnt scaling filters are symmetric, but the training loss values clearly show that only the $p = 1$ case generates a valid PRFB. The training loss for $p = 2, 3$ saturated at a high value, indicating that PR is compromised when symmetry is introduced. The $p = 1$ case results in the dilated Haar filter, because the length is greater than the minimum support required, which is in perfect agreement with Equation~\eqref{eqn:sym_orth_prfb}.}
        \label{fig:hsym_combined}
    \end{figure*}
	
	We now consider incorporating symmetry into the wavelet. 
	Symmetry results in a linear-phase response, which is important in several audio and image processing applications. If $h[n]$ is a symmetric filter, the wavelet and scaling functions obtained using the infinite cascade given in (\ref{eqn:filttowave}) would also be symmetric. Hence, it suffices to impose symmetry on the filter $\boldsymbol{h}$ to obtain a symmetric wavelet. In addition, one could also incorporate $p$ vanishing moments. Considering the factorization in \eqref{eqn:h_form_fourier} and \eqref{eqn:h_form}, and the fact that the spline component $\boldsymbol{\beta}_p$ is symmetric, it is clear that it is sufficient to impose symmetry on the trainable component $\boldsymbol{\ell}$. \\
	\indent Consider the expansion of $\boldsymbol{\ell}$ in a symmetric basis $\boldsymbol{S}$: $\boldsymbol{\ell} = \boldsymbol{S \ell^{\prime}}$, where $\boldsymbol{\ell^{\prime}}$ is the filter to be learnt. For a length-$l$ filter $\boldsymbol{h}$, and $p$ vanishing moments, $\boldsymbol{{\ell}^{\prime}} \in \mathbb{R}^{\left\lceil \frac{l - p}{2} \right\rceil}$ is the trainable component of $\boldsymbol{h}$, $\boldsymbol{\beta}_p \in \mathbb{R}^{p + 1}$ is the spline component and $\boldsymbol{S} \in \mathbb{R}^{(l - p) \times \left\lceil \frac{l - p}{2} \right\rceil}$ contains the basis vectors for length-$(l - p)$ symmetric filters. For example, for $l = 8$ and $p = 4$, $\boldsymbol{\ell^{\prime}} \in \mathbb{R}^{4}$, and
	$$
	    \boldsymbol{S} = \begin{bmatrix}
	        1 & 0 \\
	        0 & 1 \\
	        0 & 1 \\
	        1 & 0
	    \end{bmatrix}.
	$$
	The symmetric basis expansion of $\boldsymbol{\ell}$ is a sure-shot way of ensuring that the learnt scaling filter will always be symmetric. It remains to be verified whether perfect reconstruction is achieved for a chosen value of $p$.\\
	\indent The results of training symmetry-constrained filterbank autoencoders with $l = 8$ and $p$ varying from $1$ to $3$ are presented in \figurename{~\ref{fig:hsym_combined}}. The training loss vs. epochs for each case is shown on the left-most panel of \figurename{~\ref{fig:hsym_combined}}. For $p = 1$, the loss saturates at a value of $2.12 \times 10^{-26}$, with an SRER of $214.79$ dB and $\Delta_{\text{pr}} = 1.36 \times 10^{-18}$, which corresponds to a PRFB for all practical purposes. The learnt filter in this case is a circularly shifted Haar filter. The $p = 2$ and $p = 3$ cases are interesting, since the corresponding training loss values saturated at $3.73 \times 10^{-6}$ and $0.27$, respectively, with SRER $= 53.90$ dB, and $\Delta_{\text{pr}} = 0.017$ for $p = 2$, and SRER $= 5.73$ dB, and $\Delta_{\text{pr}} = 1095.06$ for $p = 3$. These results indicate that the filterbanks learnt are not PR. Therefore, the optimization has compromised PR in favour of symmetry. Despite repeating the experiments several times with different initialization, training data, various values of $p$ and $l$, we found that the learnt filterbank was not PR. These observations are completely in agreement with known results in filterbank and wavelet theory. For instance, Vaidyanathan showed that symmetric filters that satisfy the CMF condition \eqref{eqn:qmf} must have the following form (cf. Chapter 7, \cite{vaidyanathan2006multirate}):
	\begin{align}
	    \label{eqn:sym_orth_prfb}
	    h[n] = a\,\delta[n - s_1] + b\,\delta[n - s_2],
	\end{align}
	where $a, b \in \mathbb{R}$ and satisfy $a b = \frac{1}{2}$, and the shifts $s_1, s_2 \in \mathbb{Z}$ satisfy $s_1 + s_2 = l - 1$. The learnt filter in the $p = 1$ case  satisfies this requirement. Daubechies proved that the only symmetric orthogonal wavelet is the Haar wavelet \cite{daubechies1988orthonormal}. Our finding that the learnt filterbank was far from PR for $p \neq 1$ is perfectly consistent with these results.
	
	\subsection{Symmlets}
	\begin{figure*}[t]
        \centering
        \includegraphics[width=6.5in]{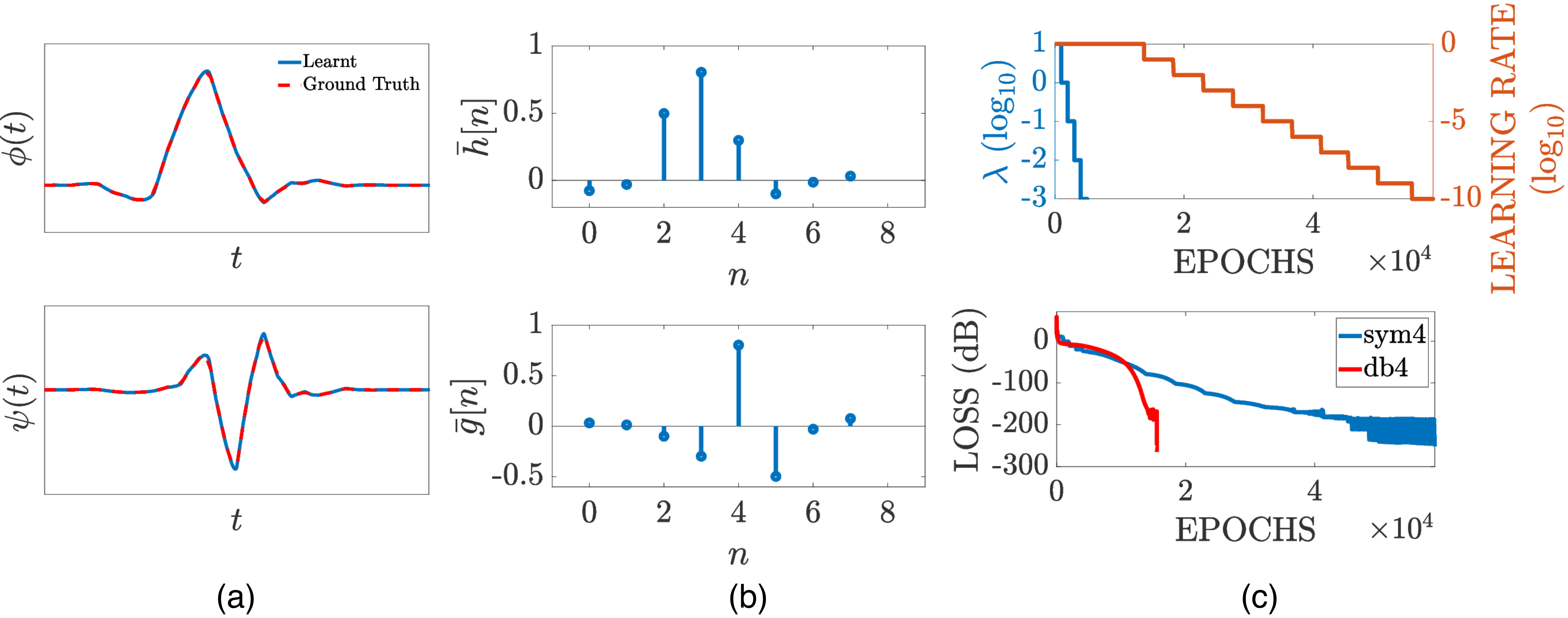}
        \caption{Learning a Symmlet: Incorporating regularization in learning a filterbank autoencoder with $\Theta_o^p$, where $l = 8$ and $p = 4$. We started with the dataset and initialization that yielded the db$4$ wavelet in \figurename{~\ref{fig:db4_repeat}}, and used a regularizer together with the MSE loss to promote symmetry about the midpoint of $h[n]$. The regularization parameter $\lambda$ was reduced in a step-wise fashion by a factor of $10$ every $1000$ iterations five times, after which it was set to $0$. The step-wise decay of $\lambda$ in the initial stages of training favours approximately symmetric solutions. As observed in the loss vs. epochs plot of \figurename{~\ref{fig:hsym_combined}}, the PR and symmetry properties are mutually exclusive, requiring that $\lambda$ be set to $0$ in order to obtain a PRFB. The learnt approximately symmetric filter turned out to be the {\it sym$4$} wavelet.}\label{fig:sym4_combined}
    \end{figure*}
	\indent If one is interested in achieving PR and also have the filters be as close to symmetric as possible, it can be done by means of a regularization penalty added to the loss function $\mathcal{L}$ instead of the symmetric basis expansion considered in Section~\ref{subsec:symm_ortho}. The regularized cost function is given by
	\begin{flalign}
	\label{eqn:sym_loss}
	    \mathcal{L}^{\text{reg}}(\mathcal{X}; \Theta_o^p) = \mathcal{L}(\mathcal{X}; \Theta_o^p) + \lambda \|\boldsymbol{h} - \boldsymbol{h}^{\text{flip}}\|_2^2,
	\end{flalign}
	where $\boldsymbol{h}^{\text{flip}}$ is the flipped version of the vector $\boldsymbol{h}$. The regularizer penalizes asymmetric filters, with the point of symmetry being the midpoint of the filter (e.g., $3.5$ for $l = 8$). We used the same dataset and initialization that yielded the db$4$ wavelet shown in \figurename{~\ref{fig:db4_repeat}} to gauge the effect of the regularizer. The regularization parameter $\lambda$ was reduced in a staircase fashion with a decay factor of $10$ for the first $5000$ iterations, after which it was set to $0$. Progressively decreasing the regularization parameter (cf. \figurename{~\ref{fig:sym4_combined}}(c)) has the effect of favouring nearly symmetric filters in the initial stages of learning, and PR in the latter stages. We found that the regularized loss $\mathcal{L}^{\text{reg}}(\mathcal{X}; \Theta_o^p)$ saturated at a value of $7.20 \times 10^{-26}$, with SRER $= 229.89$ dB, and $\Delta_{\text{pr}} = 4.19 \times 10^{-20}$, and the learnt filters generate the sym$4$ wavelet and scaling functions as shown in \figurename{~\ref{fig:sym4_combined}}(a). A similar approach can be used with choices of $l$ and $p$ other than $8$ and $4$ shown here, to obtain approximately symmetric orthogonal wavelets. Thus, the proposed wavelet learning framework allows one to generate Symmlets as well.
	
	\subsection{Learning Biorthogonal Wavelets}
	\begin{figure}[t]
		\centering
		\includegraphics[width=3.2in]{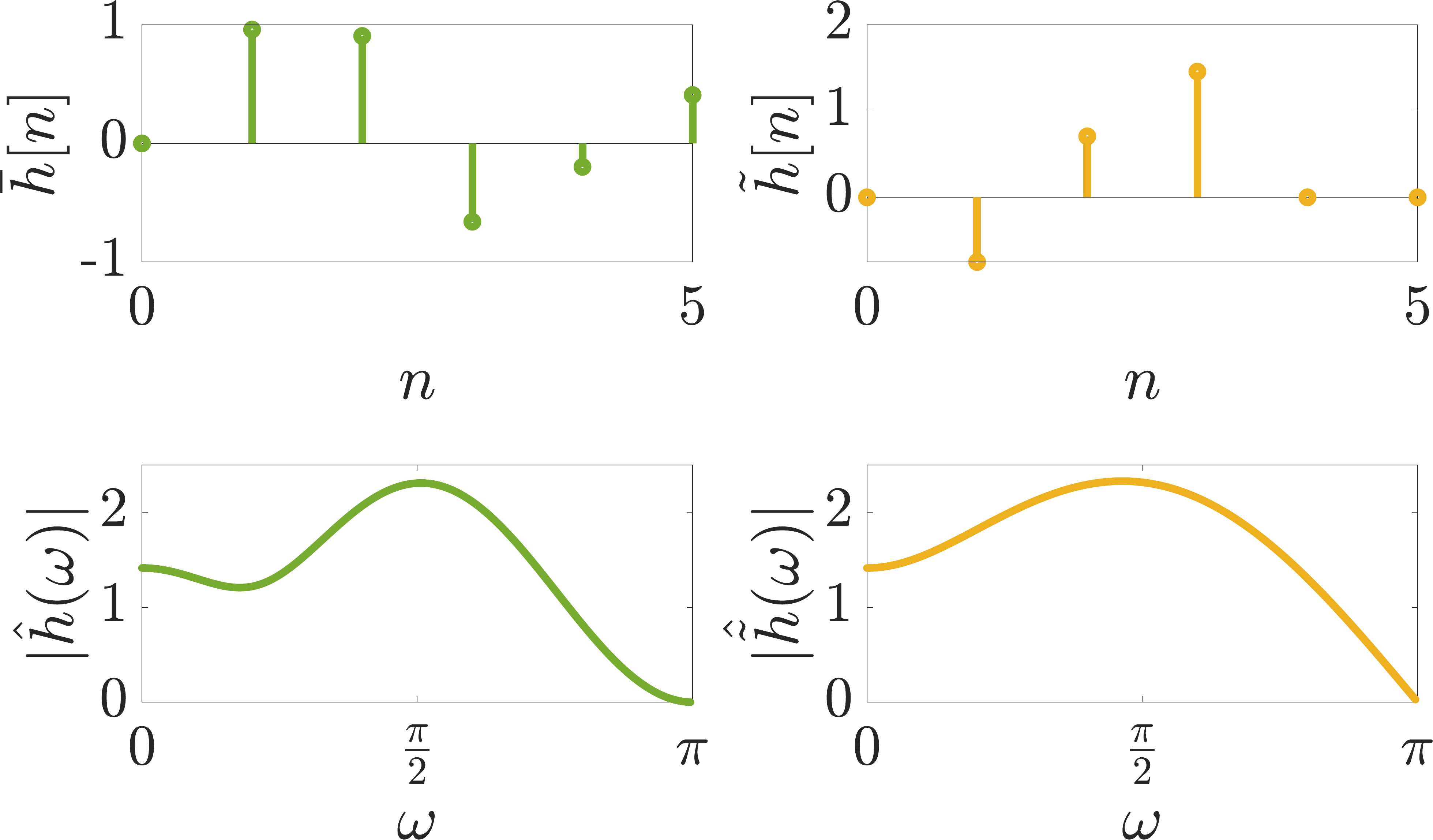}
		\caption{(Color online) A PRFB with vanishing moments that does not generate a wavelet. The scaling filters and frequency responses of a learnt filterbank autoencoder with $l = 5, \tilde{l} = 3, p=2$ and $\tilde{p} = 1$ are shown here. While these filters resulted in a PRFB, they do not generate valid scaling and wavelet functions because the infinite cascades diverge.}
		\label{fig:bior2_1}
	\end{figure}
	Biorthogonal wavelets are learnt using the autoencoder model $\Theta_b^{p, \tilde{p}}$ defined in \eqref{eqn:vc}. Unlike the orthogonal case, $\boldsymbol{\tilde{h}}$ is no longer constrained to be equal to $\boldsymbol{h}$, and therefore filter-lengths, $l$ and $\tilde{l}$, and vanishing moments, $p$ and $\tilde{p}$, must be specified. Consequently, the added flexibility allows for the construction of symmetric biorthogonal wavelets as well as wavelets with more vanishing moments than $\frac{l}{2}$. As in the orthogonal case, not all choices of the parameters yield valid wavelet generating filterbanks.\\
	\indent Invoking the factorization of $h[n]$ and $\tilde{h}[n]$ into the spline and distributional components as given in \eqref{eqn:h_form_fourier}, the PR condition \eqref{eqn:pr1} is expressed as follows:
	\begin{flalign}
	\label{eqn:bior_form}
	(1 + e^{\mathrm{j} \omega})^{p + \tilde{p}} R(\omega) + (1 - e^{\mathrm{j} \omega})^{p + \tilde{p}} R(\omega + \pi) = 2,
	\end{flalign}
	where $R(\omega) = 2^{-(p + \tilde{p})} e^{-\mathrm{j} p \omega} \hat{\ell}^*(\omega) \hat{\tilde{\ell}}(\omega)$.
	\begin{figure*}[t]
		\centering
		\includegraphics[width=6.4in]{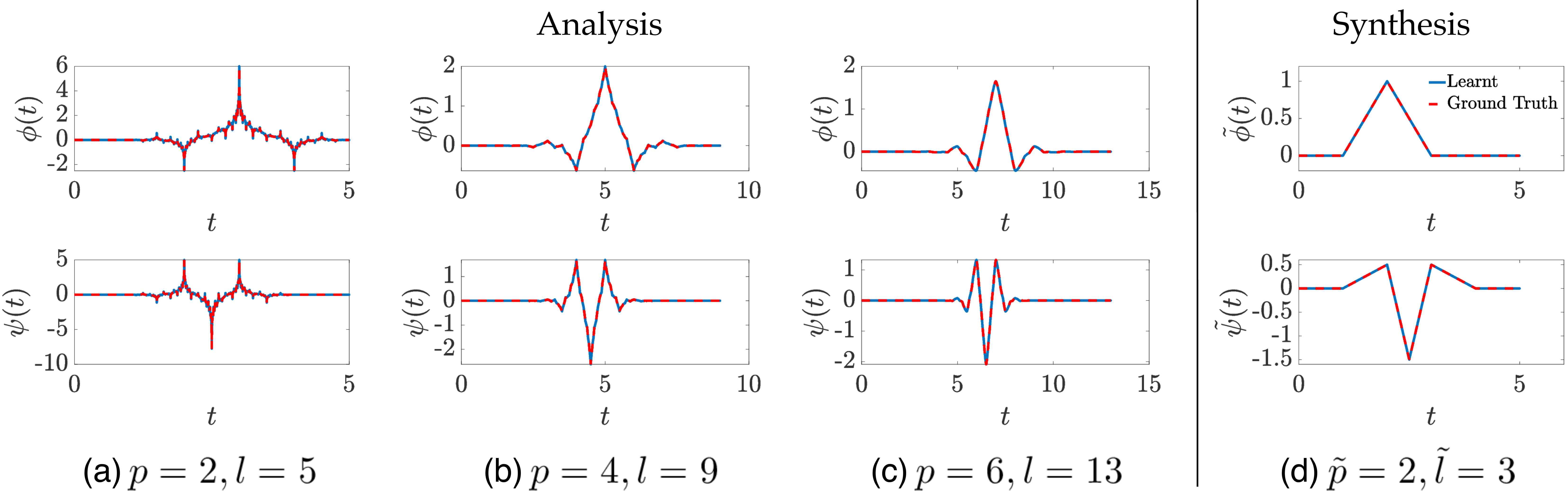}
		\caption{(Color online) Learning spline biorthogonal wavelets: The scaling and wavelet functions shown in this figure were learnt by fixing the filter $\boldsymbol{\tilde{h}}$ to the $2^{\text{nd}}$-order discrete B-spline ($\tilde{p} = 2, \tilde{l} = 3$), and setting $p = 2, 4$ and $6$, resulting in different scaling filters $\boldsymbol{h}$ having lengths $l = 5, 9,$ and $13$, respectively. Observe that the smoothness of the learnt analysis scaling and wavelet functions increases with an increase in $p$. The biorthogonal pair formed by the learnt scaling and wavelet functions in (a) and (d) correspond to the CDF 5/3 \cite{le1988sub} wavelets used in the JPEG-2000 standard for lossless compression.}
    	\label{fig:bior2_2}
	\end{figure*}
	\begin{figure*}[t]
	\centering
	\begin{tabular}{c|c}
		\subfloat[]{\includegraphics[width=3.2in]{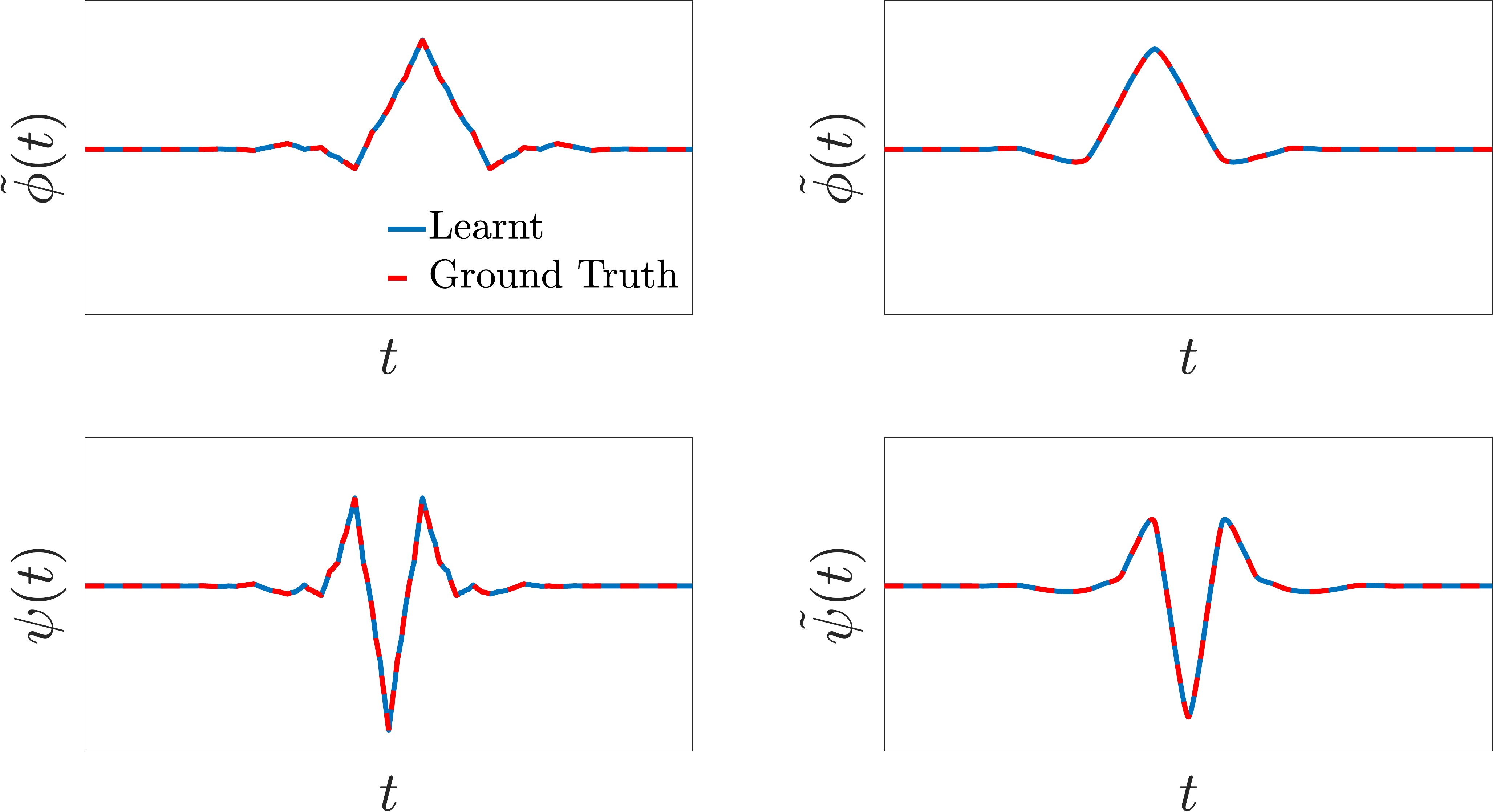}}
		&	\subfloat[]{\includegraphics[width=3.2in]{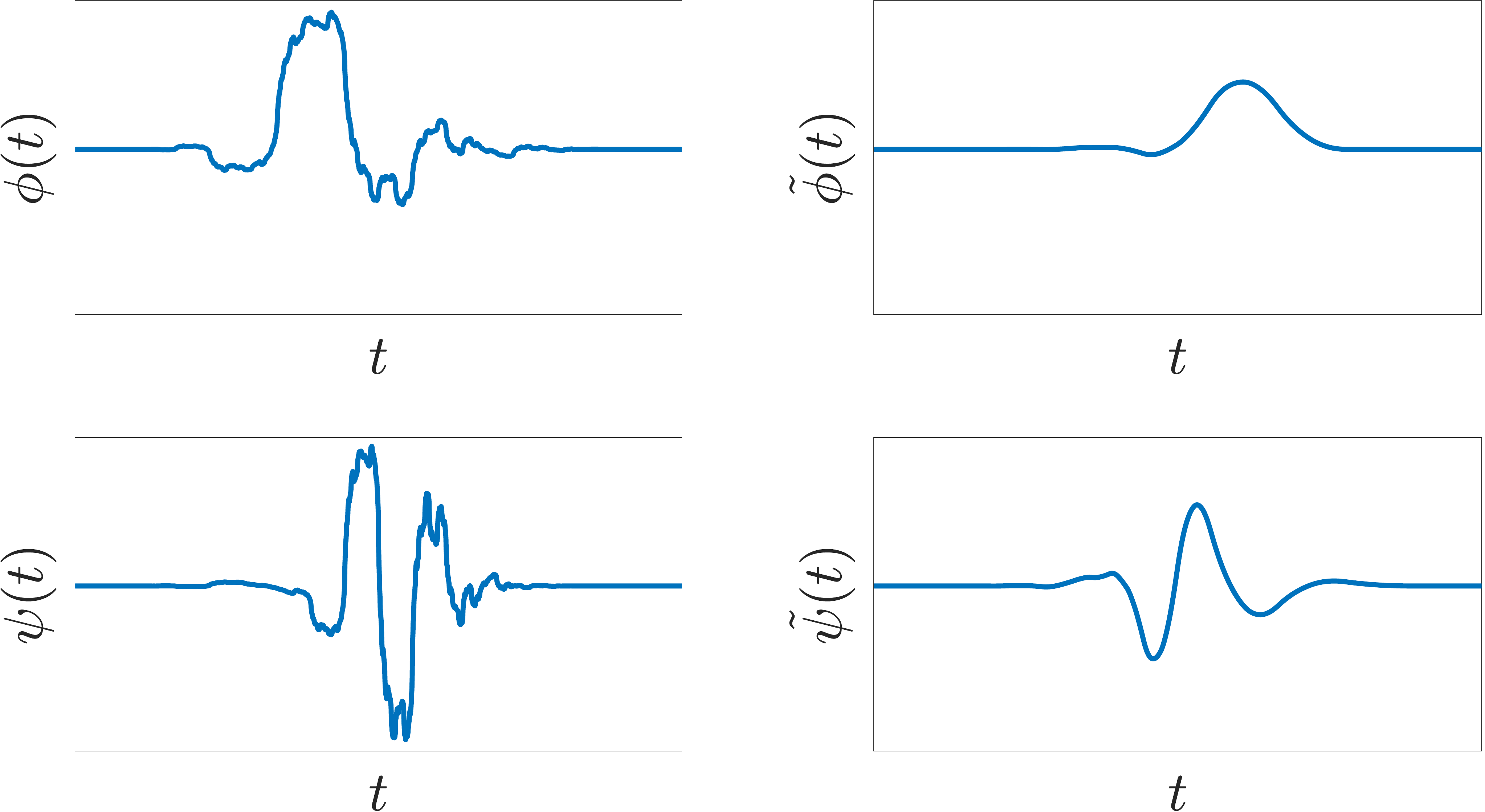}}
		\end{tabular}
		\caption{(Color online) Learnt scaling and wavelet functions for a biorthogonal autoencoder ($\Theta_b^{p, \tilde{p}}$) with vanishing moments; $l = 9, \tilde{l} = 7, p = 4$ and $\tilde{p} = 4$. (a) When the filters are constrained to be symmetric, we learn the CDF $9/7$ wavelet, used in JPEG-2000 image compression standard for lossy compression. (b) When symmetry is not enforced, new wavelet and scaling functions are learnt.}
		\label{fig:bior_4_4}
	\end{figure*}
	Once $R(\omega)$ is determined, its factorization into $\boldsymbol{\ell}$ and $\boldsymbol{\tilde{\ell}}$ determines the filters $\boldsymbol{h}$ and $\boldsymbol{\tilde{h}}$. However, not all filters that satisfy \eqref{eqn:bior_form} generate a stable wavelet basis, since the convergence of the infinite cascade is not guaranteed. One must additionally verify that the Lawton matrices \eqref{eqn:lawton} constructed using the filters $\boldsymbol{h}$ and $\boldsymbol{\tilde{h}}$ have a non-degenerate eigenspace corresponding to the eigenvalue $1$, and that the other eigenvalues are less than unity, which is a sufficient condition for convergence \cite{cohen1992stability}. An example of a learnt filterbank autoencoder that does not generate a valid wavelet biorthogonal basis is demonstrated in \figurename{~\ref{fig:bior2_1}}. The learnt filterbank had SRER of $230.69$ dB and $\Delta_{\text{pr}} = 3.49 \times 10^{-20}$, indicating that the PR condition is satisfied. However, the Lawton matrices corresponding to $\boldsymbol{h}$ and $\boldsymbol{\tilde{h}}$ were found to have three eigenvalues each with magnitudes greater than zero, indicating that the infinite cascade does not converge for either filter.\\
	\indent We now consider parameter choices corresponding to members of the Cohen-Daubechies-Feauveau (CDF) family of symmetric biorthogonal wavelets \cite{cohen1992biorthogonal}. Cohen et al. showed that, for symmetric filters $\boldsymbol{h}$ and $\boldsymbol{\tilde{h}}$ satisfying (\ref{eqn:bior_form}), the pair of vanishing moments $p$ and $\tilde{p}$, and the pair of lengths $l$ and $\tilde{l}$ must have the same parity (cf. Proposition 6.1 of \cite{cohen1992biorthogonal}), i.e., if $p$ is even, then $\tilde{p}$ must also be even; and if $p$ is odd, then $\tilde{p}$ must also be odd. The filter-lengths $l$ and $\tilde{l}$ must also follow the same pattern. Using this result to set the parameters, one could construct symmetric biorthogonal wavelets belonging to the CDF family. Two types of wavelets were constructed in \cite{cohen1992biorthogonal} using this result: spline biorthogonal wavelets and symmetric biorthogonal wavelet pairs having similar support sizes.\\
	\indent	Spline biorthogonal wavelets are obtained by setting the synthesis scaling filter $\boldsymbol{\tilde{h}}$ as the $\tilde{p}^{\text{th}}$-order discrete B-spline, and learning an appropriate analysis scaling filter with $p$ vanishing moments. Since the synthesis filter is fixed, each setting of $p$ yields a unique analysis filter satisfying the PR-$1$ condition. A higher value of $p$ yields a smoother analysis wavelet. The lengths of $\boldsymbol{\tilde{h}}$ and $\boldsymbol{h}$ are $\tilde{p} + 1$ and  $2p + \tilde{p} - 1$, respectively. Spline biorthogonal wavelet learning is illustrated in \figurename{~\ref{fig:bior2_2}}, corresponding to the choices $\tilde{p} = 2, \tilde{l} = 3$, and setting $p = 2, 4$ and $6$, resulting in a filter $\boldsymbol{h}$ of length $5, 9$ and $13$, respectively. The coefficients of the filters $\boldsymbol{h}$ and $\boldsymbol{\tilde{h}}$ were normalized to have a sum equal to $\sqrt{2}$, and the filters  $\boldsymbol{g}$ and $\boldsymbol{\tilde{g}}$ were rescaled appropriately. The learnt filterbanks had an SRER greater than $210$ dB and $\Delta_{\text{pr}}$ lesser than $3 \times 10^{-18}$. The wavelet corresponding to the choice $p = \tilde{p} = 2$, also known as the CDF 5/3 wavelet is used in the JPEG-2000 image compression scheme, and uses only integer coefficients, avoiding quantization noise and providing lossless compression. As expected, the smoothness of the learnt scaling and wavelet functions increases with an increase in the number of vanishing moments.\\
	\indent Cohen et al. also constructed another class of symmetric biorthogonal wavelets by specifying lengths $l$ and $\tilde{l}$ close to each other. For example, setting $p = 4, \tilde{p} = 4, l = 9, \tilde{l} = 7$ results in the CDF 9/7 wavelets used in the JPEG-2000 standard for lossy image compression. We considered the same set of parameters to learn a symmetric biorthogonal filterbank autoencoder. Symmetry was enforced using the symmetric basis expansion trick described in Section \ref{subsec:symm_ortho}. The learnt filterbank had an SRER of $209.95$ dB and $\Delta_{\text{pr}} = 4.21 \times 10^{-18}$, and the resulting biorthogonal scaling and wavelet functions are shown in \figurename{~\ref{fig:bior_4_4}}(a). Interestingly, for these parameter settings, regardless of the initialization and training dataset, we found that the learnt filterbank converged to the same solution. The learnt biorthogonal scaling and wavelet functions are almost identical to the CDF 9/7 counterpart. Thus, the learning approach discovered the CDF 9/7 wavelets. When the constraint of symmetry was relaxed, different solutions were obtained, which also changed with the initialization. An example of the learnt biorthogonal scaling and wavelet functions is shown in \figurename{~\ref{fig:bior_4_4}}(b), which does not exhibit symmetry. The learnt filterbank had an SRER of $212.13$ dB and $\Delta_{\text{pr}} = 2.5 \times 10^{-18}$.

\section{Conclusions}
	\label{sec:conclusions}
	We have introduced a learning approach to the problem of wavelet design, by viewing the two-channel perfect reconstruction filterbank as a convolutional autoencoder. Our approach leverages the well-established connection between perfect reconstruction filterbanks (PRFBs) and wavelets to reduce the problem to one of designing and training a filterbank autoencoder. The requirement of a certain number of {\it vanishing moments} on the wavelets is imposed by splitting the autoencoder filters into fixed and learnable parts, with the fixed part directly accounting for the vanishing moments, and the learnable part accounting for the other properties. Random Gaussian data is used to train the autoencoder by minimizing a mean-squared error (MSE) loss function. We showed that a near-zero MSE loss for a Gaussian dataset implies that the learnt filters satisfy the PR property with high probability. A learning rate schedule and stopping criteria were introduced to ensure that the learnt filterbank autoencoders are PR for all practical purposes. The training algorithm was validated by learning PRFBs with and without the vanishing moments constraint. We observed that imposing orthogonality and biorthogonality on the filterbank speeds up training considerably. This is expected because the search space of filters reduces when the constraints are imposed. Several members from the Daubechies family of minimum-phase, finite-support, orthogonal wavelets and the CDF family of symmetric biorthogonal wavelets were learnt by setting the filter lengths and vanishing moments of the filterbank autoencoder model appropriately. We also demonstrated that our framework learns wavelets outside these families. Imposing  constraints allows us to learn wavelets with the desired properties, effectively making the search for optimal wavelets a computational problem. Overall, the framework we have proposed is cogent and allows us to leverage modern machine learning tools and optimization tricks to address a problem that has been primarily approached using analytical tools.\\
	\indent Throughout this paper, we focused on 1-D wavelet learning. One could ask if the 2-D counterpart would be amenable to a similar design strategy. The answer lies in whether one is interested in designing separable or nonseparable wavelets/filterbanks. The separable 2-D design could be accomplished by a relatively straightforward extension of the 1-D learning framework. However, nonseparable wavelet design with constraints on symmetry and vanishing moments is not an easy problem. Incorporating vanishing moments using the factorization trick proposed in this paper does not carry over to 2-D. The challenge is of a fundamental nature because the roots of 2-D polynomials in general lie on curves and not points as in the 1-D case. Further, orientation plays an important role in the 2-D design aspect. Developing the 2-D counterpart of the wavelet learning framework is a fertile direction for further research.

%	\indent We are actively pursuing a 2D extension of the learning framework presented in this paper. Designing separable filterbanks, and 2D filterbanks derived from 1D PRFBs through a mapping approach based on the McClellan transform are relatively straightforward extensions of the results presented in this paper. However, designing non-separable filterbanks with control over the support dimensions and properties such as symmetry, vanishing moments, etc. offers several challenges. The roots of multidimensional polynomials lie on curves rather than points as in the 1D case, which allows one to define properties such as vanishing moments in multiple ways. It is also difficult to factorize multidimensional polynomials into component filters in the frequency domain, requiring a different approach to imposing constraints such as symmetry and vanishing moments simultaneously. The main benefit of the 2D extension would be the addition of several 2D filterbank design techniques into the repertoire of a neural network architect.

    \bibliographystyle{IEEEtran}
    \bibliography{refs.bib}

% Generated by IEEEtran.bst, version: 1.14 (2015/08/26)
\begin{thebibliography}{10}
\providecommand{\url}[1]{#1}
\csname url@samestyle\endcsname
\providecommand{\newblock}{\relax}
\providecommand{\bibinfo}[2]{#2}
\providecommand{\BIBentrySTDinterwordspacing}{\spaceskip=0pt\relax}
\providecommand{\BIBentryALTinterwordstretchfactor}{4}
\providecommand{\BIBentryALTinterwordspacing}{\spaceskip=\fontdimen2\font plus
\BIBentryALTinterwordstretchfactor\fontdimen3\font minus
  \fontdimen4\font\relax}
\providecommand{\BIBforeignlanguage}[2]{{%
\expandafter\ifx\csname l@#1\endcsname\relax
\typeout{** WARNING: IEEEtran.bst: No hyphenation pattern has been}%
\typeout{** loaded for the language `#1'. Using the pattern for}%
\typeout{** the default language instead.}%
\else
\language=\csname l@#1\endcsname
\fi
#2}}
\providecommand{\BIBdecl}{\relax}
\BIBdecl

\bibitem{icassp2019}
D.~Jawali, A.~Kumar, and C.~S. Seelamantula, ``A learning approach for wavelet
  design,'' \emph{IEEE Intl. Conf. Acoust., Speech, Signal Process.}, pp.
  5018--5022, 2019.

\bibitem{mallat1992singularity}
S.~Mallat and W.~L. Hwang, ``Singularity detection and processing with
  wavelets,'' \emph{IEEE Trans. Inf. theory}, vol.~38, no.~2, pp. 617--643,
  1992.

\bibitem{mallat1992characterization}
S.~Mallat and S.~Zhong, ``Characterization of signals from multiscale edges,''
  \emph{IEEE Trans. Patt. Anal. Mach. Intell.}, vol.~14, no.~7, pp. 710--732,
  1992.

\bibitem{donoho1995adapting}
D.~L. Donoho and I.~M. Johnstone, ``Adapting to unknown smoothness via wavelet
  shrinkage,'' \emph{J. Am. Stat. Assoc.}, vol.~90, no. 432, pp. 1200--1224,
  1995.

\bibitem{luisier2010sure}
F.~Luisier, T.~Blu, and M.~Unser, ``{SURE-LET} for orthonormal wavelet-domain
  video denoising,'' \emph{IEEE Trans. Circ. Syst. Vid. Tech.}, vol.~20, no.~6,
  pp. 913--919, 2010.

\bibitem{luisier2012cure}
F.~Luisier, T.~Blu, and P.~J. Wolfe, ``A {CURE} for noisy magnetic resonance
  images: Chi-square unbiased risk estimation,'' \emph{IEEE Trans. Image
  Process.}, vol.~21, no.~8, pp. 3454--3466, 2012.

\bibitem{seelamantula2015image}
C.~S. Seelamantula and T.~Blu, ``Image denoising in multiplicative noise,''
  \emph{IEEE Intl. Conf. Image Process.}, pp. 1528--1532, 2015.

\bibitem{figueiredo2003algorithm}
M.~A.~T. Figueiredo and R.~D. Nowak, ``An {EM} algorithm for wavelet-based
  image restoration,'' \emph{IEEE Trans. Image Process.}, vol.~12, no.~8, pp.
  906--916, 2003.

\bibitem{bioucas2007new}
J.~M. Bioucas-Dias and M.~A.~T. Figueiredo, ``A new {TwIST: Two-step Iterative
  Shrinkage/Thresholding} algorithms for image restoration,'' \emph{IEEE Trans.
  Image Process.}, vol.~16, no.~12, pp. 2992--3004, 2007.

\bibitem{li2017pure}
J.~Li, F.~Luisier, and T.~Blu, ``{PURE-LET} image deconvolution,'' \emph{IEEE
  Trans. Image Process.}, vol.~27, no.~1, pp. 92--105, 2017.

\bibitem{chan2003wavelet}
R.~H. Chan, T.~F. Chan, L.~Shen, and Z.~Shen, ``Wavelet algorithms for
  high-resolution image reconstruction,'' \emph{SIAM J. Sci. Comp.}, vol.~24,
  no.~4, pp. 1408--1432, 2003.

\bibitem{taubman2002jpeg2000}
D.~S. Taubman and M.~W. Marcellin, ``{JPEG2000}: Standard for interactive
  imaging,'' \emph{Proc. IEEE}, vol.~90, no.~8, pp. 1336--1357, 2002.

\bibitem{candes2004new}
E.~J. Cand{\`e}s and D.~L. Donoho, ``New tight frames of curvelets and optimal
  representations of objects with piecewise {$C^2$} singularities,''
  \emph{Comm. Pure Appl. Math.}, vol.~57, no.~2, pp. 219--266, 2004.

\bibitem{do2005contourlet}
M.~N. Do and M.~Vetterli, ``The contourlet transform: an efficient directional
  multiresolution image representation,'' \emph{IEEE Trans. Image Process.},
  vol.~14, no.~12, pp. 2091--2106, 2005.

\bibitem{lu2007multidimensional}
Y.~M. Lu and M.~N. Do, ``Multidimensional directional filter banks and
  surfacelets,'' \emph{IEEE Trans. Image Process.}, vol.~16, no.~4, pp.
  918--931, 2007.

\bibitem{velisavljevic2006directionlets}
V.~Velisavljevi{\`c}, B.~Beferull-Lozano, M.~Vetterli, and P.~L. Dragotti,
  ``Directionlets: Anisotropic multidirectional representation with separable
  filtering,'' \emph{IEEE Trans. Image Process.}, vol.~15, no.~7, pp.
  1916--1933, 2006.

\bibitem{jacques2011panorama}
L.~Jacques, L.~Duval, C.~Chaux, and G.~Peyr{\'e}, ``A panorama on multiscale
  geometric representations, intertwining spatial, directional and frequency
  selectivity,'' \emph{Signal Process.}, vol.~91, no.~12, pp. 2699--2730, 2011.

\bibitem{kang2017deep}
E.~Kang, J.~Min, and J.~C. Ye, ``A deep convolutional neural network using
  directional wavelets for low-dose x-ray {CT} reconstruction,'' \emph{Med.
  Phys.}, vol.~44, no.~10, pp. e360--e375, 2017.

\bibitem{liu2018multi}
P.~Liu, H.~Zhang, K.~Zhang, L.~Lin, and W.~Zuo, ``Multi-level wavelet-{CNN} for
  image restoration,'' \emph{IEEE Conf. Comp. Vis. Patt. Rec.}, pp. 773--782,
  2018.

\bibitem{bruna2013invariant}
J.~Bruna and S.~G. Mallat, ``Invariant scattering convolution networks,''
  \emph{IEEE Trans. Patt. Anal. Mach. Intell.}, vol.~35, no.~8, pp. 1872--1886,
  2013.

\bibitem{oyallon2018scattering}
E.~Oyallon, S.~Zagoruyko, G.~Huang, N.~Komodakis, S.~Lacoste-Julien,
  M.~Blaschko, and E.~Belilovsky, ``Scattering networks for hybrid
  representation learning,'' \emph{IEEE Trans. Patt. Anal. Mach. Intell.},
  vol.~41, no.~9, pp. 2208--2221, 2018.

\bibitem{kovacevic2007lifea}
J.~Kova\v{c}evi{\'c} and A.~Chebira, ``Life beyond bases: The advent of frames
  (part i),'' \emph{IEEE Signal Process. Mag.}, vol.~24, no.~4, pp. 86--104,
  2007.

\bibitem{kovacevic2007lifeb}
------, ``Life beyond bases: The advent of frames (part ii),'' \emph{IEEE
  Signal Process. Mag.}, vol.~24, no.~5, pp. 115--125, 2007.

\bibitem{primer1998introduction}
C.~S. Burrus and R.~A. Gopinath, \emph{Introduction to Wavelets and Wavelet
  Transforms}.\hskip 1em plus 0.5em minus 0.4em\relax Prentice Hall
  International, 1998.

\bibitem{mallat2008wavelet}
S.~Mallat, \emph{A Wavelet Tour of Signal Processing: The Sparse Way}.\hskip
  1em plus 0.5em minus 0.4em\relax Academic press, 2008.

\bibitem{chui2016introduction}
C.~K. Chui, \emph{An Introduction to Wavelets}.\hskip 1em plus 0.5em minus
  0.4em\relax Elsevier, 2016.

\bibitem{Vetterli86}
M.~Vetterli, ``Filter banks allowing perfect reconstruction,'' \emph{Signal
  Process.}, vol.~10, no.~3, pp. 219--244, Apr. 1986.

\bibitem{mallat1989theory}
S.~G. Mallat, ``A theory for multiresolution signal decomposition: the wavelet
  representation,'' \emph{IEEE Trans. Patt. Anal. Mach. Intell.}, vol.~11,
  no.~7, pp. 674--693, 1989.

\bibitem{daubechies1988orthonormal}
I.~Daubechies, ``Orthonormal bases of compactly supported wavelets,''
  \emph{Comm. Pure Appl. Math.}, vol.~41, no.~7, pp. 909--996, 1988.

\bibitem{daubechies1992ten}
------, \emph{Ten Lectures on Wavelets}.\hskip 1em plus 0.5em minus 0.4em\relax
  Siam, 1992, vol.~61.

\bibitem{strang2011fourier}
G.~Strang and G.~Fix, ``A fourier analysis of the finite element variational
  method,'' in \emph{Construct. Asp. Funct. Anal.}\hskip 1em plus 0.5em minus
  0.4em\relax Springer, 2011, pp. 793--840.

\bibitem{vetterli1995wavelets}
M.~Vetterli and J.~Kova\v{c}evi{\'c}, \emph{Wavelets and Subband Coding}.\hskip
  1em plus 0.5em minus 0.4em\relax Prentice-Hall, 1995.

\bibitem{strang1996wavelets}
G.~Strang and T.~Nguyen, \emph{Wavelets and Filter Banks}.\hskip 1em plus 0.5em
  minus 0.4em\relax SIAM, 1996.

\bibitem{vaidyanathan2006multirate}
P.~P. Vaidyanathan, \emph{Multirate Systems and Filter Banks}.\hskip 1em plus
  0.5em minus 0.4em\relax Pearson Education India, 2006.

\bibitem{pfister2018learning}
L.~Pfister and Y.~Bresler, ``Learning filter bank sparsifying transforms,''
  \emph{IEEE Trans. Signal Process.}, vol.~67, no.~2, pp. 504--519, 2018.

\bibitem{jmlrwaveletlearning}
C.~Tai and W.~E, ``Multiscale adaptive representation of signals: I. the basic
  framework,'' \emph{J. Mach. Learn. Res.}, vol.~17, no. 140, pp. 1--38, 2016.

\bibitem{cohen1992biorthogonal}
A.~Cohen, I.~Daubechies, and J.-C. Feauveau, ``Biorthogonal bases of compactly
  supported wavelets,'' \emph{Comm. Pure Appl. Math.}, vol.~45, no.~5, pp.
  485--560, 1992.

\bibitem{cvetkovic1998oversampled}
Z.~Cvetkovi{\`c} and M.~Vetterli, ``Oversampled filter banks,'' \emph{IEEE
  Trans. Signal Process.}, vol.~46, no.~5, pp. 1245--1255, 1998.

\bibitem{meyer1986ondelettes}
Y.~Meyer, ``Ondelettes et fonctions splines,'' \emph{S{\'e}minaire
  {\'E}quations aux d{\'e}riv{\'e}es partielles (Polytechnique)}, pp. 1--18,
  1986.

\bibitem{cohen1992stability}
A.~Cohen and I.~Daubechies, ``A stability criterion for biorthogonal wavelet
  bases and their related subband coding scheme,'' \emph{Duke Math. J.},
  vol.~68, no.~2, pp. 313--335, 1992.

\bibitem{hinton2006reducing}
G.~E. Hinton and R.~R. Salakhutdinov, ``Reducing the dimensionality of data
  with neural networks,'' \emph{Science}, vol. 313, no. 5786, pp. 504--507,
  2006.

\bibitem{ranzato2008sparse}
M.~Ranzato, Y.-L. Boureau, and Y.~L. Cun, ``Sparse feature learning for deep
  belief networks,'' in \emph{Adv. Neur. Inf. Proc. Sys.}, 2008, pp.
  1185--1192.

\bibitem{theis2017lossy}
\BIBentryALTinterwordspacing
L.~Theis, W.~Shi, A.~Cunningham, and F.~Husz{\'a}r, ``Lossy image compression
  with compressive autoencoders,'' \emph{Intl. Conf. Learn. Represent.}, 2017.
  [Online]. Available: \url{https://openreview.net/pdf?id=rJiNwv9gg}
\BIBentrySTDinterwordspacing

\bibitem{vincent2010stacked}
P.~Vincent, H.~Larochelle, I.~Lajoie, Y.~Bengio, and P.~Manzagol, ``Stacked
  denoising autoencoders: Learning useful representations in a deep network
  with a local denoising criterion,'' \emph{J. Mach. Learn. Res.}, vol.~11, no.
  Dec, pp. 3371--3408, 2010.

\bibitem{ribeiro2018study}
M.~Ribeiro, A.~E. Lazzaretti, and H.~S. Lopes, ``A study of deep convolutional
  auto-encoders for anomaly detection in videos,'' \emph{Patt. Recog. Lett.},
  vol. 105, pp. 13--22, 2018.

\bibitem{unser2003wavelet}
M.~Unser and T.~Blu, ``Wavelet theory demystified,'' \emph{IEEE Trans. Signal
  Process.}, vol.~51, no.~2, pp. 470--483, 2003.

\bibitem{unser1993ba}
M.~Unser, A.~Aldroubi, and M.~Eden, ``B-spline signal processing. {I.
  Theory},'' \emph{IEEE Trans. Signal Process.}, vol.~41, no.~2, pp. 821--833,
  1993.

\bibitem{unser1993bb}
------, ``B-spline signal processing. {II. Efficiency design and
  applications},'' \emph{IEEE Trans. Signal Process.}, vol.~41, no.~2, pp.
  834--848, 1993.

\bibitem{unser1999splines}
M.~Unser, ``Splines: A perfect fit for signal and image processing,''
  \emph{IEEE Signal Process. Mag.}, vol.~16, no.~6, pp. 22--38, 1999.

\bibitem{kingma2014adam}
D.~P. Kingma and J.~Ba, ``Adam: A method for stochastic optimization,''
  \emph{arXiv Preprint arXiv:1412.6980}, 2014.

\bibitem{blum2020foundations}
A.~Blum, J.~Hopcroft, and R.~Kannan, \emph{Foundations of Data Science}.\hskip
  1em plus 0.5em minus 0.4em\relax Cambridge University Press, 2020.

\bibitem{rudelson2013hanson}
M.~Rudelson and R.~Vershynin, ``{Hanson-Wright} inequality and sub-gaussian
  concentration,'' \emph{Electronic Communications in Probability}, vol.~18,
  2013.

\bibitem{wainwright2019high}
M.~J. Wainwright, \emph{High-dimensional Statistics: A Non-asymptotic
  Viewpoint}.\hskip 1em plus 0.5em minus 0.4em\relax Cambridge University
  Press, 2019, vol.~48.

\bibitem{boyd2004convex}
S.~Boyd and L.~Vandenberghe, \emph{Convex Optimization}.\hskip 1em plus 0.5em
  minus 0.4em\relax Cambridge University Press, 2004.

\bibitem{ieee754}
``{IEEE} standard for floating-point arithmetic,'' \emph{IEEE Std 754-2008},
  pp. 1--70, 2008.

\bibitem{tensorflow2015-whitepaper}
M.~Abadi~et al., ``Tensorflow: A system for large-scale machine learning.'' in
  \emph{OSDI}, vol.~16, 2016, pp. 265--283.

\bibitem{le1988sub}
D.~Le~Gall and A.~Tabatabai, ``Sub-band coding of digital images using
  symmetric short kernel filters and arithmetic coding techniques,'' \emph{IEEE
  Intl. Conf. Acoust., Speech, Signal Process.}, pp. 761--762, 1988.

\end{thebibliography}
    
    \begin{IEEEbiography}[{\includegraphics[width=1in,height=1.25in,clip,keepaspectratio]{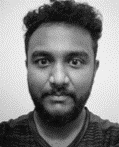}}]{Dhruv Jawali}
    received the Bachelor of Technology degree from the Department of Computer Science and Engineering, National Institute of Technology Goa, India, in 2014. He worked as a software developer at the Samsung Research Institute, Bangalore from 2014-2015. He enrolled into the PhD program at the National Mathematics Initiative, Indian Institute of Science (IISc) in August 2015, and has been working at the Spectrum Lab, Department of Electrical Engineering ever since. His research interests include wavelet theory, deep neural networks, and sparse signal processing.
    \end{IEEEbiography}
    
    \begin{IEEEbiography}[{\includegraphics[width=1in,height=1.25in,clip,keepaspectratio]{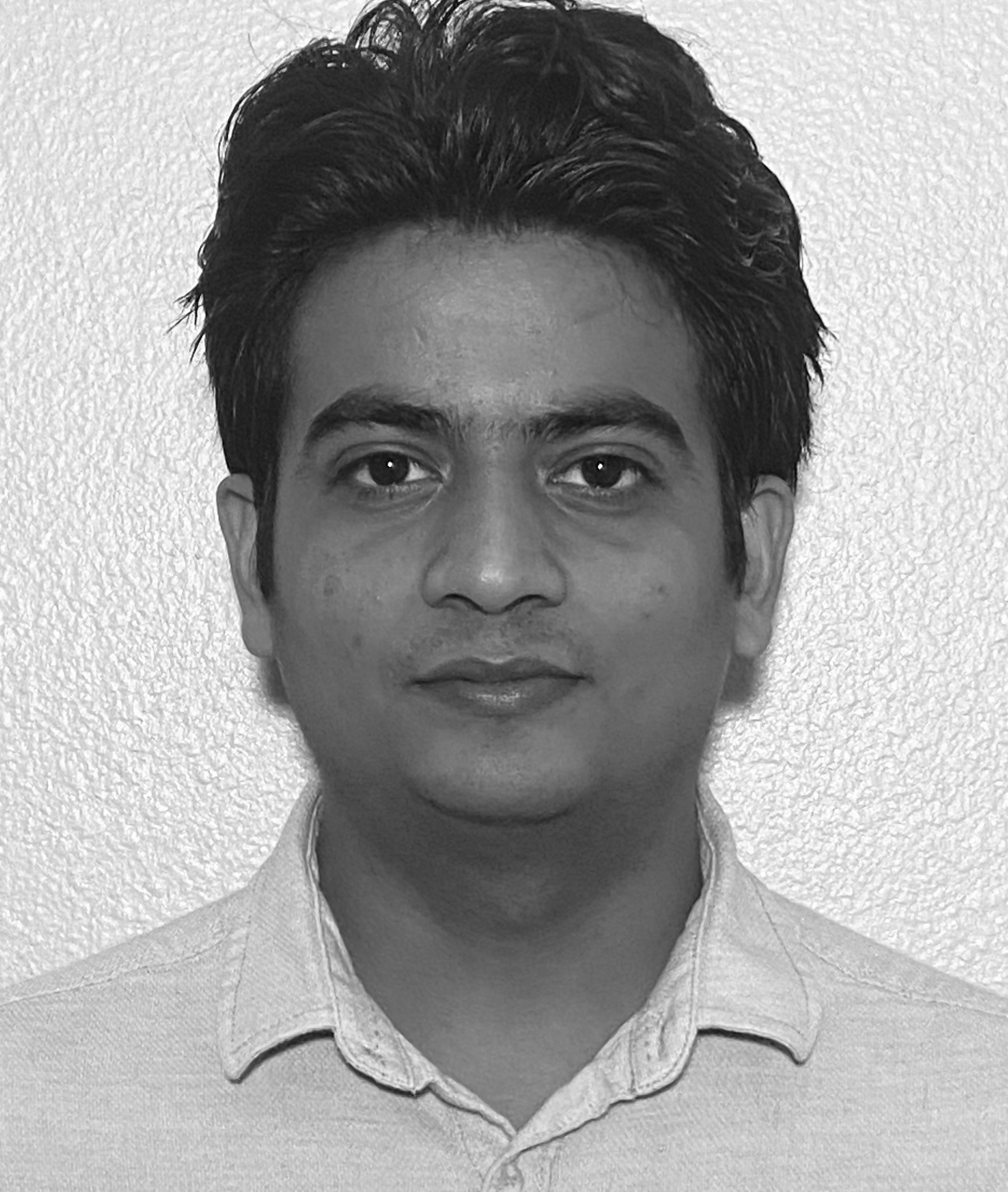}}]{Abhishek Kumar}
    received the Bachelor of Technology degree in 2014 from the Department of Electrical Engineering, Indian Institute of Technology (IIT) Indore, India. He worked as a control systems engineer in research and development division of Endurance Technologies Ltd, Aurangabad, India, from 2014 to 2016. He received the Masters in Artificial intelligence degree from Indian Institute of science (IISc), Bangalore, India, in 2019. He is currently a graduate student in the Department of Electrical and Computer Engineering at Rice University, Houston, USA. His research interest include signal processing, wireless networks, deep learning on graphs and reinforcement learning.     
    \end{IEEEbiography}
    
    % insert where needed to balance the two columns on the last page with
    % biographies
    %\newpage
    
    \begin{IEEEbiography}[{\includegraphics[width=1in,height=1.25in,clip,keepaspectratio]{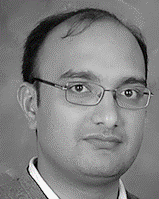}}]{Chandra Sekhar Seelamantula}
    (M'99--SM'13) received the Bachelor of Engineering degree in 1999 with a Gold Medal and the Best Thesis Award from the Osmania University College of Engineering, India, with a specialization in Electronics and Communication Engineering. He received the Ph.D. degree from the Department of Electrical Communication Engineering, Indian Institute of Science (IISc), Bangalore, in 2005. During April 2005 -- March 2006, he was a Technology Consultant for M/s. ESQUBE Communication Solutions Private Limited, Bangalore, and developed proprietary audio coding solutions. During April 2006 -- July 2009, he was a Postdoctoral Fellow in the Biomedical Imaging Group, Ecole Polytechnique F\'ed\'erale de Lausanne (EPFL), Switzerland, where he specialized in the fields of biomedical imaging, splines, wavelets, and sampling theories. In July 2009, he joined the Department of Electrical Engineering, IISc., where he is currently an Associate Professor and directs the Spectrum Lab. He is also an Associate Faculty in the Centre for Neuroscience, IISc. He is currently the Chair of the IEEE Signal Processing Society Bangalore Chapter, a Senior Area Editor of the IEEE Signal Processing Letters, an Associate Editor of IEEE Transactions on Image Processing, a member of the IEEE Technical Committee on Computational Imaging, and an Area Chair of IEEE International Conference on Acoustics, Speech, and Signal Processing 2021. He received the Prof. Priti Shankar Teaching Award from IISc in 2013. His research interests include signal processing, sampling theory, inverse problems in computational imaging, and machine learning.
    \end{IEEEbiography}
\appendices
\section{Proof of Proposition 1}
    \label{sec:appendix_form}
    \setcounter{prop}{0}
    \begin{prop}
	    \label{prop:loss_form_app}
	    For a set of filters $$\Theta= \left\{ \boldsymbol{h}, \boldsymbol{\tilde{g}} \in \mathbb{R}^{l}, \boldsymbol{\tilde{h}}, \boldsymbol{g} \in \mathbb{R}^{\tilde{l}} \right\},$$ and a dataset $\mathcal{X}= \{ \boldsymbol{x}_j \in \mathbb{R}^{s} \}_{j = 1}^{m}$, there exists a matrix $\boldsymbol{B} \in \mathbb{R}^{s \times s}$ such that the loss function defined in Equation \textsc{(21)} in the main document can be expressed as
	    \begin{align}
	        \label{eqn:loss_form_app}
	        \mathcal{L}(\mathcal{X}; \Theta) = \frac{1}{m}\sum_{j = 1}^{m} \|\boldsymbol{B} \boldsymbol{x}_j\|_2^2.
	    \end{align}
	    If the filters form a PRFB, then $\boldsymbol{B} = \boldsymbol{0}_s$ and vice versa.
	\end{prop}
    \begin{IEEEproof}
        The linear convolution of $\boldsymbol{x} \in \mathbb{R}^{s}$ with $\boldsymbol{h} \in \mathbb{R}^{l}$ may be expressed as the matrix operation $\boldsymbol{H} \boldsymbol{x}$,  where $\boldsymbol{H} \in \mathbb{R}^{(l + s - 1) \times s}$ is a Toeplitz matrix constructed from $\boldsymbol{h}$.
        % The cascade of the downsampling and upsampling blocks results in the odd-indexed entries of the input vector becoming $0$, and may be represented as $\boldsymbol{D} \boldsymbol{x}$, where $\boldsymbol{D} \in \mathbb{R}^{s \times s}$ is a diagonal matrix with entries $\boldsymbol{D}_{i, i} = 1$ for $i$ even, and $\boldsymbol{D}_{i, i} = 0$ for $i$ odd.
        If $\tilde{l} < l$, we pad the filters of length $\tilde{l}$ with $\left\lfloor \frac{l - \tilde{l}}{2} \right\rfloor$ zeros on the left and $\left\lceil \frac{l - \tilde{l}}{2} \right\rceil$ on the right, and vice versa. In both cases, the net delay caused by the filterbank is $\tau - 1$, where $\tau = \max(l, \tilde{l})$. Effectively, the output of the filterbank can be expressed as follows:
        \begin{align*}
        \boldsymbol{\hat{x}}_j^{\prime}(\Theta) = \left(\boldsymbol{\tilde{H}} \boldsymbol{D} \boldsymbol{H} + \boldsymbol{\tilde{G}} \boldsymbol{D} \boldsymbol{G} \right) \boldsymbol{x}_j,
        \end{align*}
        where $\boldsymbol{H}, \boldsymbol{G} \in \mathbb{R}^{(s + \tau - 1) \times s}$ and $\boldsymbol{\tilde{H}}, \boldsymbol{\tilde{G}} \in \mathbb{R}^{(s + 2 \tau - 2) \times (s + \tau - 1)}$ are linear convolution matrices corresponding to $\boldsymbol{h}, \boldsymbol{g}, \boldsymbol{\tilde{h}}$ and $\boldsymbol{\tilde{g}}$, respectively, and $\boldsymbol{D} \in \mathbb{R}^{(s + \tau - 1) \times (s + \tau - 1)}$ represents the cascade of the downsample and upsample operations. The cascade results in the odd-indexed entries of the input becoming $0$, while the even-indexed entries remain unchanged. Hence, $\boldsymbol{D}$ is a diagonal matrix, with entries $\boldsymbol{D}_{i, i} = 1$ for $i$ even, and $\boldsymbol{D}_{i, i} = 0$ for $i$ odd. The dimensions of $\boldsymbol{\hat{x}}_j^{\prime}(\Theta)$ do not match with that of $\boldsymbol{x}_j$ because of the linear convolution operation. Therefore, the loss is computed considering a truncated version of $\boldsymbol{\hat{x}}_j^{\prime}(\Theta)$, denoted as $\boldsymbol{\hat{x}}_j(\Theta)$. More precisely, since the overall delay of the filterbank is $\tau - 1$, the middle portion of $\boldsymbol{\hat{x}}_j^{\prime}(\Theta)$ is extracted by pre-multiplying it with a matrix $\boldsymbol{P} \in \mathbb{R}^{s \times s + 2 \tau - 2}$ defined as
        $
        \boldsymbol{P} = \begin{bmatrix} \boldsymbol{Z}_{\tau - 1},
        \boldsymbol{I}_{s},  \boldsymbol{Z}_{\tau - 1}
        \end{bmatrix}
        $, where $\boldsymbol{Z}_{\tau - 1}$ is a zero matrix of dimensions $s \times \tau - 1$. The preceding considerations lead to the following expression for the loss $$\mathcal{L}(\mathcal{X}; \Theta) = \frac{1}{m}\sum_{j = 1}^{m} \|\boldsymbol{B} \boldsymbol{x}_j\|_2^2,$$
        where $$\boldsymbol{B} = \boldsymbol{I}_s - \boldsymbol{P} \left(\boldsymbol{\tilde{H}} \boldsymbol{D} \boldsymbol{H} + \boldsymbol{\tilde{G}} \boldsymbol{D} \boldsymbol{G} \right).$$ When $\boldsymbol{B} = \boldsymbol{0},$ perfect reconstruction is achieved because the net transfer function of the filterbank autoencoder equals the identity. Consequently, $$ \boldsymbol{P} \left(\boldsymbol{\tilde{H}} \boldsymbol{D} \boldsymbol{H} + \boldsymbol{\tilde{G}} \boldsymbol{D} \boldsymbol{G} \right) = \boldsymbol{I}_s.$$ The diagonal entries of $\boldsymbol{B}$ correspond to the PR-1 conditions and the off-diagonal ones correspond to the PR-2 conditions.
    \end{IEEEproof}

\section{Proof of Theorem \ref{thm:conc_loss_app}}
    \label{sec:appendix_conc}
    \setcounter{theorem}{2}
    \begin{theorem}
        \label{thm:conc_loss_app}
        For the loss function $\mathcal{L}$ defined in \eqref{eqn:loss_form_app}, where $\mathcal{X}$ comprises standard Gaussian vectors, the deviation of the loss from its expected value is bounded in probability as follows:
        \begin{align*}
        \mathbb{P}(\big| \mathcal{L} - \mathbb{E} \{ \mathcal{L} \} \big| \geq k) \leq \begin{cases} 
                2 e^{\frac{-k^2 s}{8 \|\boldsymbol{B}\|_{2}^{2} \|\boldsymbol{B}\|_{\textsc{F}}^{2}}}, \, &k \leq \|\boldsymbol{B}\|_{\textsc{F}}^2, \\
                2 e^{\frac{-k s}{8 \|\boldsymbol{B}\|_{2}^{2}}}, \, &k > \|\boldsymbol{B}\|_{\textsc{F}}^2.
            \end{cases}
    \end{align*}
    \end{theorem}
    \begin{IEEEproof}
        Here, we provide the key results together with their proofs that ultimately lead to the proof of Theorem \ref{thm:conc_loss_app}. To begin with, we analyze the concentration of $\boldsymbol{Bx}$ for $\boldsymbol{B} \in \mathbb{R}^{s \times s}$ and $\boldsymbol{x} \sim \mathcal{N}(\boldsymbol{0}, \boldsymbol{I}_s)$.
        \begin{lemma}
        	\label{thm:gen_annulus}
        	Let $\boldsymbol{x} \sim \mathcal{N}(\boldsymbol{0}, \boldsymbol{I}_s)$ and $\boldsymbol{B} \in \mathbb{R}^{s \times s}$ be a deterministic matrix. Then, for $k \geq 0$,
        	\begin{align*}
        	\mathbb{P}\left( \big| \| \boldsymbol{B} \boldsymbol{x} \|_2^2 - \|\boldsymbol{B}\|_{\textsc{F}}^2 
        	\big| \geq  k \right) \leq \begin{cases} 
        	2 e^{\frac{-k^2}{8 \|\boldsymbol{B}\|_{2}^{2} \|\boldsymbol{B}\|_{\textsc{F}}^{2}}}, &k \leq \|\boldsymbol{B}\|_{\textsc{F}}^2, \\
        	2 e^{\frac{-k}{8 \|\boldsymbol{B}\|_{2}^{2}}}, &k > \|\boldsymbol{B}\|_{\textsc{F}}^2.
        	\end{cases}
        	\end{align*}
        \end{lemma}
        \begin{IEEEproof}
        	Let $X:= \|\boldsymbol{B} \boldsymbol{x} \|_2^2 - \|\boldsymbol{B}\|_{\text{F}}^2$. The moment generating function (m.g.f.) of $X$ is
        	\begin{align*}
        	M_{X}(t) = \mathbb{E}\left\{ e^{t X} \right\} = \mathbb{E} \left\{ e^{t (\boldsymbol{x}^\textsc{T}\boldsymbol{B}^\textsc{T}\boldsymbol{B}\boldsymbol{x} - \|\boldsymbol{B}\|_{\text{F}}^2 )} \right\}.
        	\end{align*}
        	The spectral decomposition of $\boldsymbol{B}^\textsc{T}\boldsymbol{B}$ is written as  $$\boldsymbol{B}^\text{T}\boldsymbol{B} =  \boldsymbol{U} \Lambda \boldsymbol{U}^{\text{T}} = \sum_{i = 1}^{s} \lambda_i \boldsymbol{u}_i \boldsymbol{u}_i^{\text{T}},$$
        	where $\lambda_i$ and $\boldsymbol{u}_i$ are the corresponding $i^{\text{th}}$ eigenvalue and eigenvector, respectively, and $\boldsymbol{U}$ is a unitary matrix with the eigenvectors as its columns. Let $\sigma_i$ denote the $i^{\text{th}}$ singular value of $\boldsymbol{B}$. Since $\sigma_{i}^2 = \lambda_i$, and $\|\boldsymbol{B}\|_{\text{F}}^2 = \sum_{i = 1}^{s} \sigma_i^2$,
        	\begin{align*}
        	M_{X}(t) = \mathbb{E} \left\{ e^{t \sum_{i = 1}^{s} \left[ \sigma_i^2 (\boldsymbol{u}_i^{\text{T}} \boldsymbol{x} )^2 - \sigma_i^2 \right] } \right\}.
        	\end{align*}
        	The vector $\boldsymbol{U}^{\text{T}} \boldsymbol{x} = \left[ \boldsymbol{u}_1^{\text{T}} \boldsymbol{x}, \ldots, \boldsymbol{u}_s^{\text{T}} \boldsymbol{x} \right]^{\textsc{T}}$ has the same distribution as $\boldsymbol{x}$ due to the rotation-invariance/unitary-invariance property of $\mathcal{N}(\boldsymbol{0}, \boldsymbol{I}_s)$. Therefore, $z_i := \boldsymbol{u}_i^{\text{T}} \boldsymbol{x}$, $i = 1, \ldots, s$, are independent and identically distributed Gaussian random variables with mean $0$ and variance $1$. Hence,
        	\begin{align*}
        	M_{X}(t) &= \mathbb{E} \left\{ e^{\sum_{i = 1}^{s} t \sigma_i^2 \left(  z_i^2 - 1 \right)} \right\} = \prod_{i = 1}^{s} \mathbb{E} \left\{ e^{t \sigma_i^2 \left( z_i^2 - 1 \right)} \right\} \\
        	&= \prod_{i = 1}^{s} \frac{1}{\sqrt{2 \pi}} \int\limits_{-\infty}^{+\infty} e^{t \sigma_i^2 \left( z_i^2 - 1 \right) } \cdot e^{-\frac{z_i^2}{2}} \mathrm{d}z_i \\
        	&= \prod_{i = 1}^{s} \frac{e^{-t \sigma_i^2}}{\sqrt{2 \pi}} \int\limits_{-\infty}^{+\infty} e^{-\left(1 - 2 t \sigma_i^2  \right) \frac{z_i^2}{2}} \mathrm{d}z_i\\
        	&= \prod_{i = 1}^{s} \frac{e^{-t \sigma_i^2}}{(1 - 2 t \sigma_i^2)^{\frac{1}{2}}}, \qquad t \leq \frac{1}{2 \sigma_i^2}, i = 1, \ldots, s.
        	\end{align*}
        	We now establish that $X$ is a sub-exponential random variable by bounding the m.g.f., which is facilitated by the following inequality (Chapter 2, Example $2.8$ of \cite{wainwright2019high}):
        	\begin{align}
        	\label{eqn:up_bound}
        	\frac{e^{-x}}{(1 - 2x)^{\frac{1}{2}}} \leq e^{2 x^2}, \quad |x| \leq \frac{1}{4}.
        	\end{align}
        	The proof of \eqref{eqn:up_bound} is given in Appendix \ref{sec:proof_ineq}.
        	%        Let $x_i = t \sigma_i^2$, and define $f(x_i) = e^{-x_i} (1 - 2x_i)^{-\frac{1}{2}}$. To derive probability bounds, it is sufficient to upper-bound the moment generating function $M_{X}(t)$ in a neighbourhood of $0$. Using \ref{lem:up_bound} to upper-bound each $f(x_i)$, for $1 \leq i \leq s$. 
        	\noindent Hence,
        	\begin{align*}
        	M_{X}(t) \leq \prod_{i = 1}^{s} e^{2 t^2 \sigma_i^4} = e^{2 t^2 \sum\limits_{i = 1}^{s} \sigma_i^4}, \quad \text{ for } |t| \leq \frac{1}{4 \|\boldsymbol{B}\|_2^2}.
        	\end{align*}
        	Note that $\sum_{i = 1}^{s} \sigma_i^4 = \sum_{i = 1}^{s} \lambda_i^2 = \|\boldsymbol{B}^{\text{T}} \boldsymbol{B}\|_{\text{F}}^2$, and
        	\begin{align*}
        	\|\boldsymbol{B}^{\text{T}} \boldsymbol{B}\|_{\text{F}}^2 = \sum\limits_{i = 1}^{s} \| \boldsymbol{B}^{\text{T}} \boldsymbol{b}_i \|_2^2 \leq \sum_{i = 1}^{s} \|\boldsymbol{B}^{\text{T}}\|_2^2 \|\boldsymbol{b}_i \|_2^2 = \|\boldsymbol{B}\|_2^2 \|\boldsymbol{B}\|_{\text{F}}^2,
        	\end{align*}
        	where $\boldsymbol{b}_i$ is the $i^{\text{th}}$ column of $\boldsymbol{B}$, and the inequality follows from the definition of the matrix $2$-norm. Hence,
        	\begin{align}
        	\label{eqn:mgf_upper_bound}
        	M_{X}(t) \leq e^{2 t^2 \|\boldsymbol{B}\|_2^2 \|\boldsymbol{B}\|_{\text{F}}^2}, \quad \text{ for } |t| \leq \frac{1}{4 \|\boldsymbol{B}\|_2^2}.
        	\end{align}
        	The tail bounds are obtained from the m.g.f. using the Chernoff approach. For $t \geq 0$ and $k \geq 0$,
        	\begin{align*}
        	\mathbb{P}(X \geq k) = \mathbb{P}(e^{t X} \geq e^{t k}) \leq \frac{\mathbb{E}\{e^{t X} \}  }{e^{t k}} = \frac{M_{X}(t)}{e^{t k}},
        	\end{align*}
        	using Markov's inequality. The Chernoff bound is obtained by minimizing $e^{-t k} M_{X}(t)$ with respect to $t \in \left(0, \frac{1}{4 \|\boldsymbol{B}\|_2^2} \right)$:
        	\begin{align*}
        	\log \mathbb{P}(X \geq k) &\leq \min_{t} \left\{ \log M_{X}(t) - t k\right\}\\
        	&\leq \min_{t} \left\{ 2 t^2 \|\boldsymbol{B}\|_2^2 \|\boldsymbol{B}\|_{\text{F}}^2 - t k \right\}.
        	\end{align*}
        	The minimum value of the parabola $g(t, k) := 2 t^2 \|\boldsymbol{B}\|_2^2 \|\boldsymbol{B}\|_{\text{F}}^2 - t k$ occurs at $t= \frac{k}{4 \|\boldsymbol{B}\|_2^2 \|\boldsymbol{B}\|_{\text{F}}^2}$. If $k \leq \|\boldsymbol{B}\|_{\text{F}}^2$, then $t \leq \frac{1}{4 \|\boldsymbol{B}\|_2^2}$, and the minimum in the interval $t \in \left(0, \frac{1}{4 \|\boldsymbol{B}\|_2^2} \right)$ is the global minimum. Hence
        	\begin{align}
        	\label{eqn:upper_dev_1}
        	\mathbb{P}(X \geq k) &\leq e^{-\frac{k^2}{8 \|\boldsymbol{B}\|_2^2 \|\boldsymbol{B}\|_{\text{F}}^2}}, \quad \text{ for } k \leq \|\boldsymbol{B}\|_{\text{F}}^2.
        	\end{align}
        	If $k > \|\boldsymbol{B}\|_{\text{F}}^2$, then $t > \frac{1}{4 \|\boldsymbol{B}\|_2^2}$, and the minimum in the interval occurs at the boundary point $t = \frac{1}{4 \|\boldsymbol{B}\|_2^2}$, since $g(t, k)$ is a monotonically decreasing function in the interval $t \in \left(0, \frac{1}{4 \|\boldsymbol{B}\|_2^2} \right)$, and for $k > \|\boldsymbol{B}\|_{\text{F}}^2$ we have
        	\begin{eqnarray}
        	\log \mathbb{P}(X \geq k) &\leq \frac{\|\boldsymbol{B}\|_{\text{F}}^2}{8 \|\boldsymbol{B}\|_2^2} - \frac{k}{4 \|\boldsymbol{B}\|_2^2} \leq \frac{-k}{8 \|\boldsymbol{B}\|_2^2}\nonumber\\
        	\Rightarrow  \mathbb{P}(X \geq k) &\leq e^{\frac{-k}{8 \|\boldsymbol{B}\|_2^2}} \quad \text{ for } k > \|\boldsymbol{B}\|_{\text{F}}^2.
        	\label{eqn:upper_dev_2}
        	\end{eqnarray}
        	
        	To obtain the lower deviates, let $Y = -X = \|\boldsymbol{B}\|_{\text{F}}^2 - \|\boldsymbol{B} \boldsymbol{x} \|_2^2$, and note that $M_Y(t)$ is only defined for $t > -\frac{1}{2 \|\boldsymbol{B}\|_{2}^2}$. In the interval $(-\frac{1}{2}, \frac{1}{2})$, $M_Y(t) = M_X(-t)$, and hence, the inequality  in \eqref{eqn:up_bound} could be used to upper-bound $M_Y(t)$. Thus,
        	\begin{align*}
        	M_{Y}(t) \leq e^{2 t^2 \|\boldsymbol{B}\|_2^2 \|\boldsymbol{B}\|_{\text{F}}^2}, \quad \text{ for } |t| \leq \frac{1}{4 \|\boldsymbol{B}\|_2^2},
        	\end{align*}
        	which is similar to \eqref{eqn:mgf_upper_bound}. The Chernoff bounds for $Y$ are the same as those for $X$, and $\mathbb{P}(Y \geq k) = \mathbb{P}(X \leq -k)$. Hence,
        	\begin{align}
        	\label{eqn:lower_dev}
        	\mathbb{P}(X \leq -k) &\leq \begin{cases}
        	e^{-\frac{k^2}{8 \|\boldsymbol{B}\|_2^2 \|\boldsymbol{B}\|_{\text{F}}^2}}, \quad &k \leq \|\boldsymbol{B}\|_{\text{F}}^2, \\
        	e^{\frac{-k}{8 \|\boldsymbol{B}\|_2^2}}, \quad &k > \|\boldsymbol{B}\|_{\text{F}}^2.
        	\end{cases}
        	\end{align}
        	Consolidating \eqref{eqn:upper_dev_1}, \eqref{eqn:upper_dev_2}, and \eqref{eqn:lower_dev}, and invoking the \emph{union bound}, we have the required result.
        \end{IEEEproof}
        Having established the concentration of $\boldsymbol{Bx}$, we next establish similar concentration inequalities for the loss function $\mathcal{L}(\mathcal{X}; \Theta) = \frac{1}{s} \sum_{i = 1}^{s} \|\boldsymbol{B} \boldsymbol{x_i} \|_2^2$, where $\boldsymbol{x}_i  \sim \mathcal{N}(\boldsymbol{0}, \boldsymbol{I}_s)$ are independent and identically distributed. Let $X_i:= \|\boldsymbol{B} \boldsymbol{x_i} \|_2^2 - \|\boldsymbol{B}\|_{\text{F}}^2$. The m.g.f. of the centered loss function
        $$\mathcal{L}_c := \mathcal{L}(\mathcal{X}; \Theta) - \mathbb{E} \{ \mathcal{L}(\mathcal{X}; \Theta) \} = \left(\frac{1}{s} \sum_{i = 1}^{s} \|\boldsymbol{B} \boldsymbol{x_i} \|_2^2 \right) - \|\boldsymbol{B}\|_{\text{F}}^2$$
        is given by 
        \begin{align*}
        M_{\mathcal{L}_c}(t) &= \mathbb{E} \left\{ e^{t \mathcal{L}_c} \right\} = \mathbb{E} \left\{ e^{\frac{t}{s} \sum_{i = 1}^{s} X_i} \right\}\\
        &= \prod_{i = 1}^s \mathbb{E} \left\{ e^{\frac{t}{s} X_i} \right\} = \prod_{i = 1}^{s} M_{X_i}\left( \frac{t}{s} \right).
        \end{align*}
        Invoking the inequality on the m.g.f. in \eqref{eqn:mgf_upper_bound}, we obtain
        \begin{align*}
        M_{\mathcal{L}_c}(t) \leq e^{2 \frac{t^2}{s} \|\boldsymbol{B}\|_{2}^2 \|\boldsymbol{B}\|_{\text{F}}^2}, \quad \text{ for } t \leq \frac{s}{4 \|\boldsymbol{B}\|_{2}^2}.
        \end{align*}
        Hence, the Chernoff bound for the upper deviate is given by
        \begin{align*}
        \mathbb{P}(\mathcal{L}_c \geq k) &\leq \begin{cases}
        e^{-\frac{k^2 s}{8 \|\boldsymbol{B}\|_2^2 \|\boldsymbol{B}\|_{\text{F}}^2}}, \quad &k \leq \|\boldsymbol{B}\|_{\text{F}}^2, \\
        e^{\frac{-k s}{8 \|\boldsymbol{B}\|_2^2}}, \quad &k > \|\boldsymbol{B}\|_{\text{F}}^2.
        \end{cases}
        \end{align*}
        Similarly, considering $- \mathcal{L}_c$ instead of $\mathcal{L}_c$ gives us the bound for the lower deviate. Combining the two, we get
        \begin{align*}
        \mathbb{P}(\big| \mathcal{L}_c \big| \geq k) \leq \begin{cases} 
        2 e^{\frac{-k^2 s}{8 \|\boldsymbol{B}\|_{2}^{2} \|\boldsymbol{B}\|_{\textsc{F}}^{2}}}, \, &k \leq \|\boldsymbol{B}\|_{\textsc{F}}^2, \\
        2 e^{\frac{-k s}{8 \|\boldsymbol{B}\|_{2}^{2}}}, \, &k > \|\boldsymbol{B}\|_{\textsc{F}}^2.
        \end{cases}
        \end{align*}
        Thus, Theorem \ref{thm:conc_loss_app} stands proved.
    \end{IEEEproof}

\section{Proof of the inequality in \eqref{eqn:up_bound}}
    \label{sec:proof_ineq}
    Let $f(x) = e^{-x} (1 - 2x)^{-\frac{1}{2}}$ and $ g(x) =  e^{2 x^2}$. Then, $f^{\prime}(x) = 2x e^{-x} (1 - 2x)^{-\frac{3}{2}}$ and $ g^{\prime}(x) =  4 x e^{2 x^2}$. Let
    $$
    h(x) = \frac{g^{\prime}(x)}{f^{\prime}(x)} = 2 e^{2x^2 + x} (1 - 2x)^{\frac{3}{2}}.
    $$
    For $x < \frac{1}{2}$, $h(x)$ is strictly positive. The derivative of $h(x)$ is
    $$
    h^{\prime}(x) = 2 e^{2x^2 + x} (1 - 2x)^{\frac{1}{2}} \left( - 8 x^2 + 2x - 2 \right).
    $$
    The factor $\left( - 8 x^2 + 2x - 2 \right)$ is strictly negative for all values of $x$. Hence, $h^{\prime}(x)$ is strictly negative for $x < \frac{1}{2}$, and therefore $h(x)$ is a monotonically decreasing in the interval $(-\infty, \frac{1}{2})$. Consequently, for $x \leq \frac{1}{4}$, $h(x) \geq h(\frac{1}{4}) \geq 1$. Hence, for all $|x| \leq \frac{1}{4}$, $h(x) \geq 1$. Further, we note that both $f^{\prime}(x)$ and $g^{\prime}(x)$ have the same sign. Combining this property with $h(x) \geq 1$, gives $
    |g^{\prime}(x)| \geq |f^{\prime}(x)|.$
    Since $f(0) = g(0) = 1$, integrating over the intervals $\left[-\frac{1}{4}, 0\right]$ and $\left[0, \frac{1}{4}\right]$, we obtain that $g(x) \geq f(x)$ for $x \in \left[-\frac{1}{4}, \frac{1}{4}\right]$. \hfill\IEEEQEDopen
\end{document}